\documentclass[5p]{elsarticle}

\usepackage{hyperref}
\usepackage{amsmath, cases}
\usepackage{multirow}

\usepackage{algorithm}
\usepackage{algorithmic}

\biboptions{authoryear}

\begin{document}

\begin{frontmatter}

\title{Towards Understanding Chinese Checkers with Heuristics, Monte Carlo Tree Search, and Deep Reinforcement Learning}

\author[address]{Ziyu Liu\fnref{fn1}}
\ead{zliu6676@uni.sydney.edu.au}

\author[address]{Meng Zhou\fnref{fn1}}
\ead{mzho7212@uni.sydney.edu.au}

\author[address]{Weiqing Cao\corref{correspondingauthor}\fnref{fn1}}
\ead{wcao4942@uni.sydney.edu.au}

\author[address]{Qiang Qu}
\ead{qiqu7152@uni.sydney.edu.au}

\author[address]{Henry Wing Fung Yeung}
\ead{hyeu8081@uni.sydney.edu.au}

\author[address]{Vera Yuk Ying Chung}
\ead{vera.chung@sydney.edu.au}

\cortext[correspondingauthor]{Corresponding author}
\fntext[fn1]{Equal contribution}
\address[address]{School of Computer Science, The University of Sydney, Camperdown NSW 2006, Australia}

\begin{abstract}
The game of Chinese Checkers is a challenging traditional board game of perfect information for computer programs to tackle that differs from other traditional games in two main aspects: first, unlike Chess, all checkers remain indefinitely in the game and hence the branching factor of the search tree does not decrease as the game progresses; second, unlike Go, there are also no upper bounds on the depth of the search tree since repetitions and backward movements are allowed. In this work, we present an approach that effectively combines the use of heuristics, Monte Carlo tree search, and deep reinforcement learning for building a Chinese Checkers agent without the use of any human game-play data. In addition, unlike other common approaches, our approach uses a two-stage training pipeline to facilitate agent convergence. Experiment results show that our agent is competent under different scenarios and reaches the level of experienced human players.
\end{abstract}


\end{frontmatter}

\section{Introduction}
There have been many successes \citep{ref1, ref2, ref3, ref24} in developing machine learning algorithms that excel at traditional zero-sum board games of perfect information, such as Chess \citep{ref1, thrun1995learning} and Go \citep{ref2}, as well as other games \citep{nipsmarkovgames, aaaizerosumgame, icmlzerosumgame, pendharkar2012game, guo2014deep, stanley2006real, finnsson2008simulation, chen2017robust}. Little research attention, however, has been drawn to solve the game of Chinese Checkers with machine learning techniques. While there are known strategies for approaching Chinese Checkers, such strategies often only focus on the initial starting policy and locally optimal game-play patterns, or that they may heavily rely on cooperation between the players, which is often not possible. In addition, with the goal to move all of a player’s checkers to the opponent’s side, there are two particular aspects of the game that are different from other traditional games which may lead to an enormously large game-tree and state-space complexity \citep{ref7} and game divergence. First, checkers in the game remain indefinitely on the board and cannot be captured; and second, the possibility of repetition and backward movements of checkers mean that the game can be arbitrarily long without violating game rules.
 
In this work, we present an approach that effectively combines the use of heuristics, Monte Carlo Tree Search (MCTS), and reinforcement learning for building a Chinese Checkers agent without the use of any human game-play data. Our approach is inspired by AlphaGo Zero on the game of Go \citep{ref2}, where the agent is represented as a single, deep residual \citep{ref27} convolutional neural network \citep{lenet} which takes as input the current game state and outputs both the current game-play policy and the current game state value from the perspective of the current player. The training pipeline for the agent, however, differs in that the agent is first guided and initialized using a greedy heuristic, and experiment results show that such guidance can quickly allow the agent to learn very basic strategies like consecutive hops and can significantly reduce the depth of the search trees in the games. The next stage of the training pipeline aims to improve the network through reinforcement learning, where each iteration the network generates many self-play games with Monte Carlo Tree Search and is trained using the final game outcomes and post-search action policies. For comparison, we also trained a network entirely through MCTS-guided self-play reinforcement learning (i.e. \textit{tabula rasa} learning) as well as a network trained with the Deep Q-Learning approach as seen in the work of \cite{ref30} used for training agents to play Atari Games. Experiment results show that the current approach outperforms both of these approaches to a significant extent.
 
The rest of the paper is organised as follows. Section 2 discusses related work done in the field. Section 3 explains in detail the methods and techniques used for building the agent. Section 4 presents and extensively discusses the results of the experiments conducted on several important aspects of the training framework and discusses how they may affect the agent’s performance. Section 5 concludes the paper.

\section{Background}
\paragraph{Heuristics in games} Heuristics are one of the most widely and commonly used techniques in artificial intelligence to guide decision-making in games when no exact \textit{solutions} are available \citep{oh2017playing}, or the search spaces are too large for any form of exhaustive search \citep{ref14}.  With a carefully designed heuristic function for game state transitions, the branching factor of the search tree can be significantly reduced (as evidently bad moves can be avoided) while potentially increasing the performance of the agent. Different forms of heuristic search have been successfully applied and shown to improve the performance of an agent in various games such as the work of \cite{ref8}, \cite{ref12}, \cite{ref14}, \cite{ref15}, \cite{ref16}, and \cite{ref17}. This includes various board games such as Chess and Go \citep{ref16, ref17} where heuristic search can be used for finding locally optimal plays and various video games such as path-finding through complex terrain and performing complex real-time strategy \citep{ref8, ref12, ref14, ref15}. In the case of Chinese checkers, heuristics has also been applied, for example, in search for the quickest game progress by prioritising moving the most forward checkers, or in search for the shortest possible game \citep{ref5}. However, while these heuristic rules can inspire some game-play strategies like building “bridges” (depicted in Figure 4), they depend heavily on the state of the game and sometimes the cooperation of the opponent, which means achieving expert-level play solely with heuristics is often not possible in real and competitive games.

\paragraph{Monte Carlo Tree Search} Originally introduced in the work of \cite{ref18}, Monte Carlo tree search (MCTS) is an effective method for searching the next optimal action when given the current game state \citep{ref2, ref3, ref17, ref21, ref22, ref23, ref24}. It does so by running multiple \textit{simulations}, where each simulation gradually expands the search-tree rooted at the current game state by following the locally optimal action at each subsequent state while taking into account some degree of exploration through to a leaf node in the search-tree and “back-up” the value of the leaf state for the next simulation \citep{ref18}; an action is then decided based on all simulations. With its effective look-ahead search, MCTS has been shown to improve the performance of agents in games such as Ms Pac-Man \citep{ref23}, and board games such as Go (both in restricted \citep{ref17} and unrestricted games \citep{ref2, ref3}), the Settlers of Catan \citep{ref22} and Kriegspiel \citep{ref21}. Modifications to MCTS is also possible; for example, \cite{ref24} demonstrated in their work that by taking into account human and game-specific factors, MCTS can be modified to further improve the agent’s performance. In our work, a slightly modified version (as compared to common variants in as seen in \cite{ref17, ref18, ref19, ref21}) of MCTS is used, where we replaced the Upper Confidence Bound (UCB) algorithm with a variant as used in AlphaGo Zero \citep{ref2}. More details are presented in Methods.

\paragraph{Reinforcement Learning in games} In recent years, reinforcement learning has been shown to be the key in the successes of many intelligent agents excelling at board games that were thought to be extremely hard or even impossible for computers to master, including Go, Chess, Othello \citep{ref1, ref2, ref3, ref25}, and others. It is a powerful framework that allows computer programs to learn through interactions with the environment rather than following a fixed set of rules or learning from a fixed set of data, and it allows the agent to be more flexible with the complex, dynamic real-world interactions, hence suitable for games in particular. In the context of board games, the reinforcement learning framework can also be combined with various other techniques such as Monte Carlo tree search to further its effectiveness. For example, AlphaZero \citep{ref33} was able to achieve expert-level performance in Chess, Go, and Shogi without the use of any human knowledge when trained solely on self-play reinforcement learning guided by Monte Carlo tree search. In our work, reinforcement learning is used to continually improve the performance of our agent through MCTS-guided self-plays.

\paragraph{Deep Q-Learning} As a well-known reinforcement learning approach, Q-learning follows the same structure as the Markov Decision Process to maximize the accumulated immediate reward obtained by performing a specific action for a given state. For a long time since its development, neural networks has been abandoned for this approach due to its lack of convergence guarantee. However, firstly proposed by \cite{ref30}, Deep Q-Learning with the use of deep neural networks has been successfully applied to Atari games to outperforms human in particular games, as well as various other games and applications \citep{van2016deep, greenwald2003correlated, tesauro2004extending, ref31, ref32}. By using some training techniques such as experience replay, epsilon-greedy and Huber loss, the convergence rate of the network is significantly increased. In our case, a Deep Q-Learning agent is trained following the algorithm and training techniques proposed by \cite{ref30} as a comparison to our main approach. The details will be discussed in Methods.

\section{Methods}
\subsection{Game Formulation}
In a standard game of Chinese Checkers \citep{ref4}, there can be 2, 3, 4, or 6 players playing on a star-shaped board, each with 10 checkers. In this work, we reduced the game for streamlined training and analysis: we first restricted the number of players to 2 and the number of checkers for each player to 6, and then resized to game board to a 49-slot, diamond-shaped board by removing the extra checker slots for the extra players. All game rules are unaffected. Figure 1 illustrates the revised game board as adapted from a standard 6-player board. To represent the board in a program, we use a \(7\times7\) matrix, called the “\textit{game matrix}”, in which the value 0 represent an empty slot, value 1 represent checkers of the first player (Player 1) and value 2 for the second player (Player 2). Player 1 always starts from the bottom-left corner (i.e. bottom of the board) while Player 2 always starts from the top-right corner (i.e. top of the board) of the game matrix. Figure 2 illustrates a mapping between the board representation used in this paper and its game matrix representation. Win state in the game can be easily checked by looking only at the top-right and bottom-left corners of the game matrix.

\begin{figure}[t]
    \centering
    \includegraphics[width=0.4\textwidth]{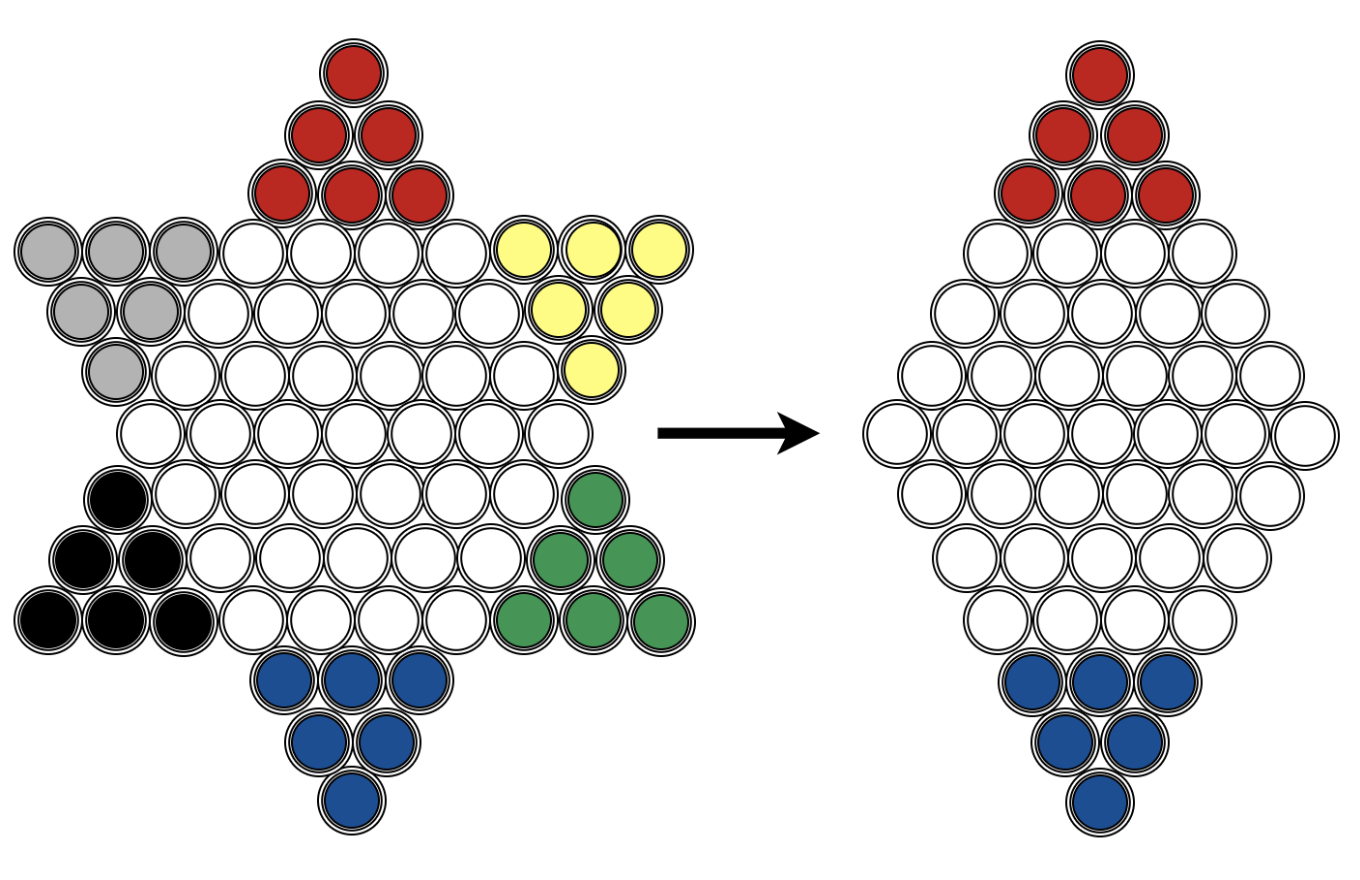}
    \caption{Restricted game board for two players. White cells represent empty slots and coloured cells represent slots occupied by different players.}
    \label{fig:fig1}
\end{figure}

\begin{figure}[t]
    \centering
    \includegraphics[width=0.4\textwidth]{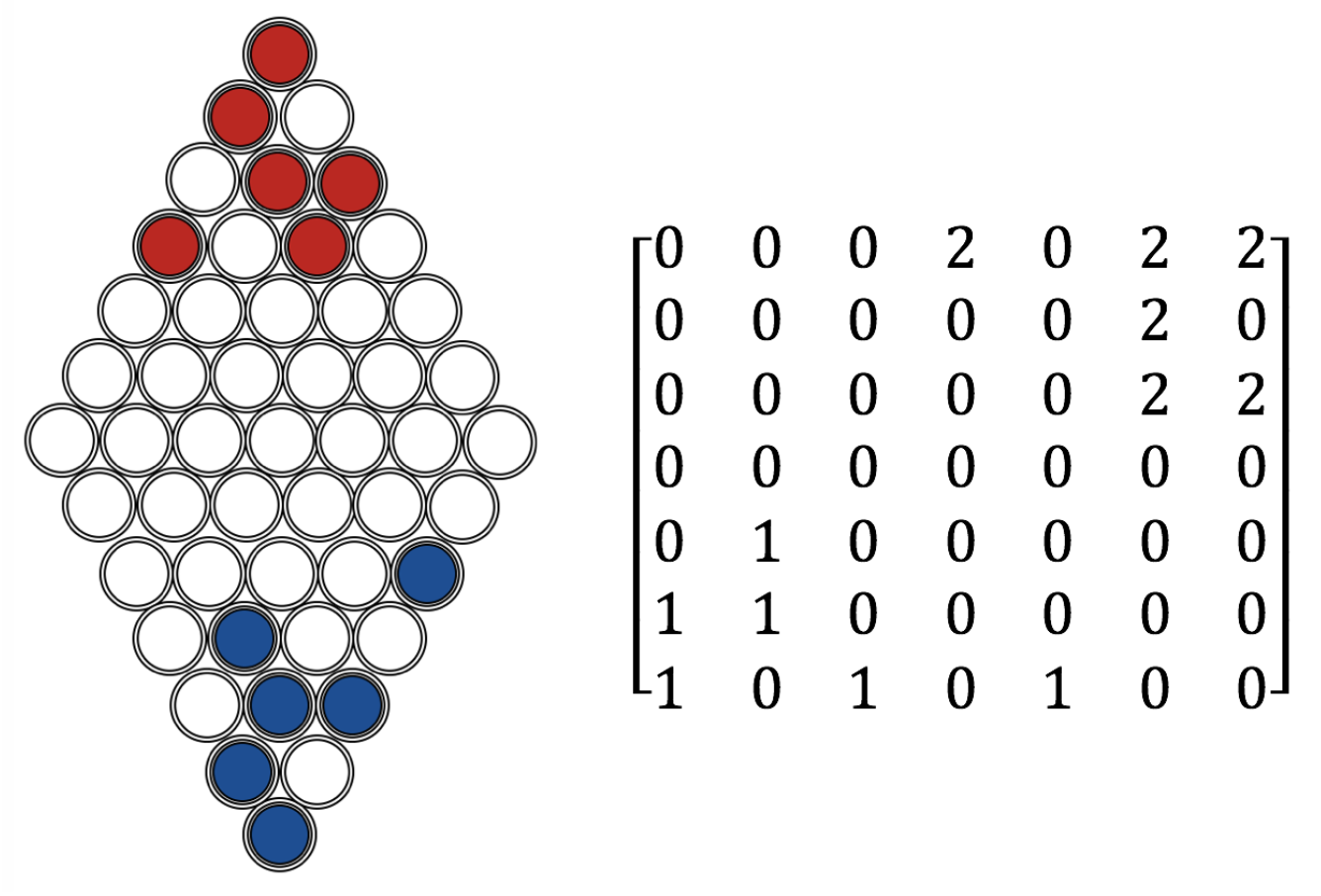}
    \caption{Matrix representation of a game state.}
    \label{fig:fig2}
\end{figure}

\begin{figure}[t]
    \centering
    \includegraphics[width=0.3\textwidth]{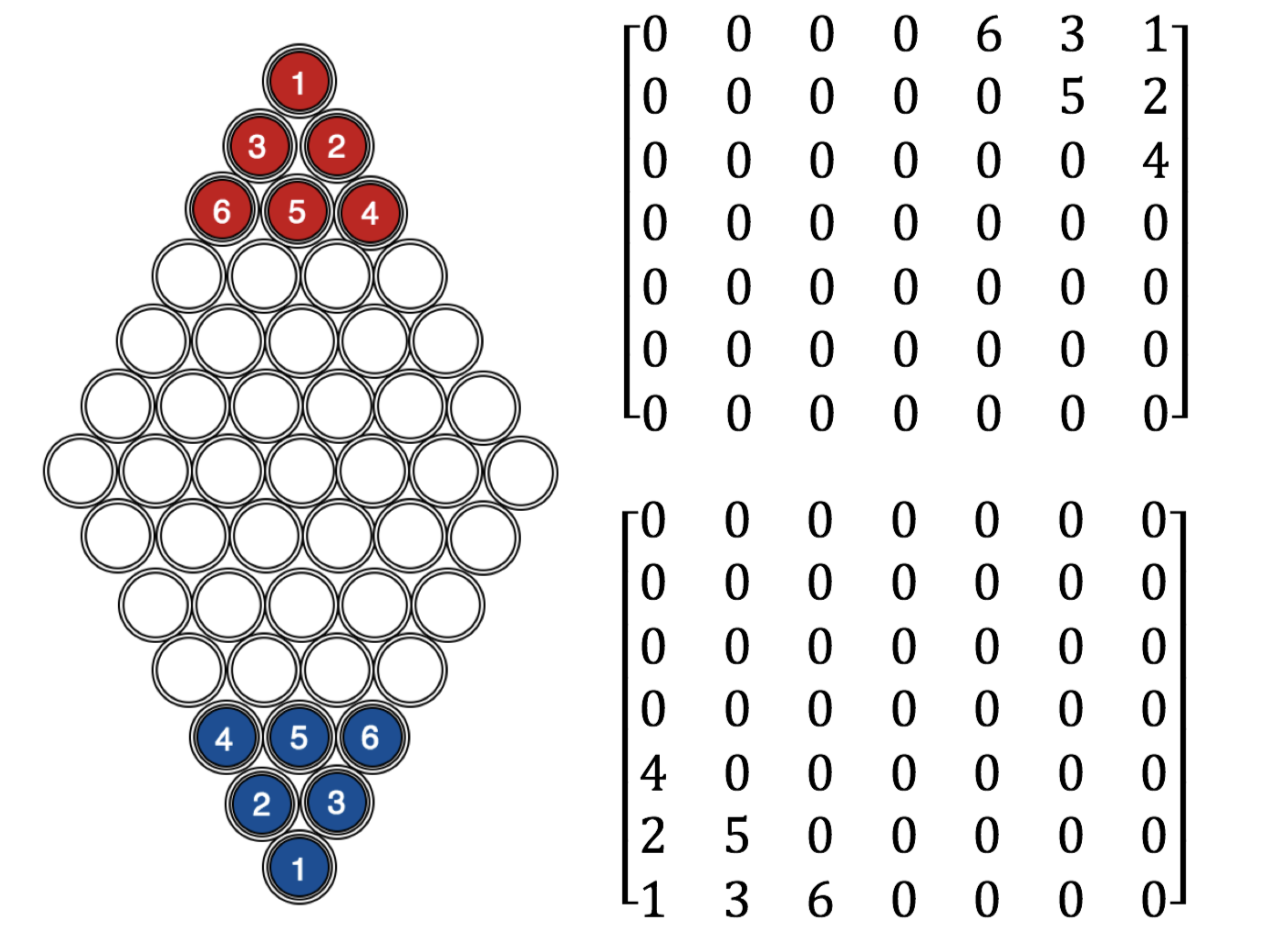}
    \caption{ID matrices of a single game state as input to neural network. The IDs written on the checkers (1 to 6) correspond to values in the matrices. The top and bottom matrices are produced from the perspective of Player 1 (blue) and Player 2 (red) respectively.}
    \label{fig:fig3}
\end{figure}

Each checker is tracked throughout a game by maintaining its ID and its position as the game progresses. The position of a checker is its position in the game matrix, and its ID is an integer between 1 to 6 as illustrated in Figure 3. Four lookup tables, from position to ID and from ID to position for each player, are cached for performance and are updated as each move is made. The reason for maintaining an ID for each checker is that when the agent makes a prediction given a game state, a unique identifier for each checker is required such that the agent’s predicted action policy is not ambiguous.

Due to the nature of the game that a player can take arbitrarily many hops for moving a checker to a single destination, valid moves for a player at any given game state are represented as a collection of distinct starting-ending position pairs. The valid moves for each player is determined by first checking the adjacent slots for each checker which are reachable by “rolling”. Then, for each checker of the current player, the valid destinations are determined through recursive mirror “hops” through depth-first search. By using the unique IDs of the checkers, we can therefore represent an agent’s action policy, which is a vector output from the network representing the prior probabilities for each possible move, at any given state using an ID-and-destination mapping instead of a from-and-to mapping, thereby greatly reducing the output dimension. For this reason, the game matrix is primarily used for storage and visualisation purposes, while inputs to the agent will take the form of the “\textit{ID matrices}” as illustrated in Figure 3. Further details on the input and output of the agent will be discussed in the Network Architecture and Configurations subsection.

\begin{figure}[t]
    \centering
    \includegraphics[width=0.4\textwidth]{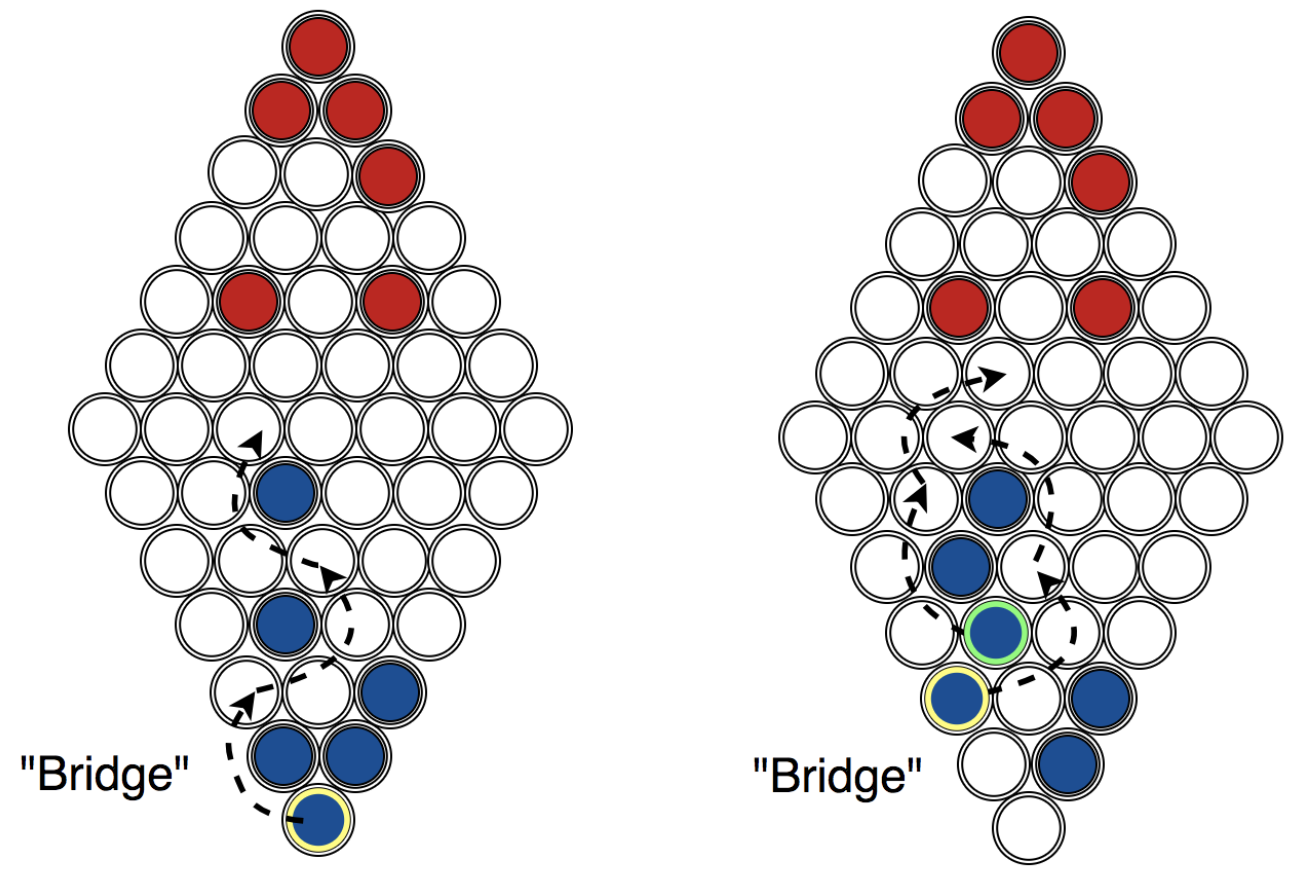}
    \caption{A local game-play strategy used in Chinese Checkers, commonly referred as a “bridge”. In the figure on the right, the yellow highlighted checker first makes two consecutive hops, then another “bridge” is formed for green-highlighted checker in the next round.}
    \label{fig:fig4}
\end{figure} 

\begin{figure}[t]
    \centering
    \includegraphics[width=0.4\textwidth]{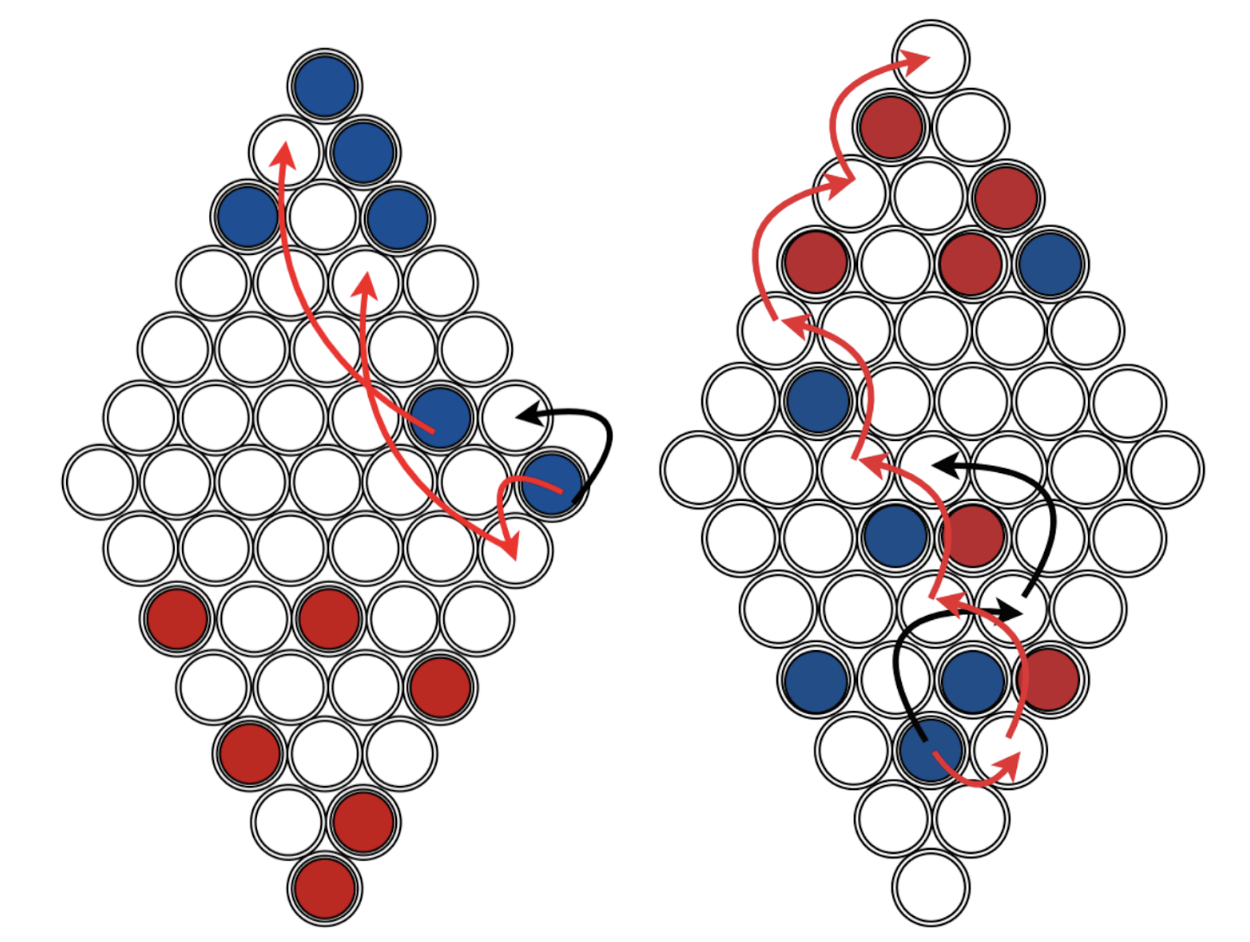}
    \caption{Examples of sub-optimal plays (black arrow) by heuristic guided agent vs optimal plays (red arrows).}
    \label{fig:fig5}
\end{figure}

In addition to maintaining checker positions we also cache a fixed number of past game states as history. Past game states are particularly useful for two reasons. First, some common game-play strategies (such as ``Bridge'' as shown in Figure 4) are not directly observable for the current state while such strategies are being formed. It may therefore be useful for the agent to look at a longer history before making a move. Secondly, game history can be used for checking meaningless repetitions or cyclic moves during training by examining the number of distinct destinations of past moves. Specifically, if in the past 16 moves of the game there are less than 6 unique move destinations, then it can be reasonably inferred that at least one player is performing move cycles of length smaller than or equal to 3. However, note that it remains a difficult task to detect cyclic moves of arbitrary length. In our work, 16 nearest past games states are kept for detecting short cyclic moves, and 3 nearest past game states were used together as input to model.

\subsection{Heuristic}
There are two aspects of Chinese Checkers that make it challenging for an agent to perform a deep search down the game tree. First, unlike Chess, no capturing of checkers is possible in Chinese Checkers, which means that the number of checkers in the game (and hence the branching factor of the search tree) does not decrease as the game progresses. Second, unlike Go, repetitions and backward movements are allowed, which means there are also no upper bounds on the depth of the search tree. Even in our restricted game instance where the average length of rational games is 45 moves, there are still on average 30 possible moves at each state (both statistics calculated from a large number of plays), yielding a total of more than \(10^{66}\) unique move sequences.

In order to effectively reduce both the breadth and depth of any form of search, we propose a simple heuristic: at any given state, the value of the next possible valid actions (determined as described in subsection Game Formulation) are ranked in decreasing order by their \textit{forward distance}, which is denoted by \( \Psi(A) \) for an action \(A\). Denote Player 1 and Player 2 to be \(P1\) and \(P2\) respectively, and the action \( A = (A_s, A_e) \), where \( A_s = (A_{sa}, A_{sb}) \) and \( A_e = (A_{ea}, A_{eb}) \) refer to the 2-tuple of starting position \( A_s \) and ending position \( A_e \) of the checker move in terms of their row, column coordinates in the game matrix respectively. Then, for any valid \( A_s \) and \( A_e \) where \( 1 \leq A_{sa} \leq 7 \), \( 1 \leq A_{sb} \leq 7 \), \( 1 \leq A_{ea} \leq 7  \), and \( 1 \leq A_{eb} \leq 7 \) since the game matrix has 7 rows and columns, \( \Psi(A) \) is given by:

\begin{numcases}{\Psi(A)=}
  (A_{sa}-A_{sb})-(A_{ea}-A_{eb}), & for \(P1\) \\
  (A_{ea}-A_{eb})-(A_{sa}-A_{sb}), & for \(P2\)
\end{numcases} \\

Intuitively, if we consider the actual game board instead of the game matrix, then \( \Psi(A) \) can be understood as how far the action \(A\) has moved the checker in the positive direction, which is upwards for Player 1 and downwards for Player 2 on the board. With \( \Psi \), we can then formulate two greedy agents, being stochastically greedy and deterministically greedy respectively: the former uses the value \(\frac{\Psi(A_i)}{\sum_{j}^{} \Psi(A_j)}\) as the prior probability for selecting the action \(A_i\) among all possible actions \(A_j\), while the deterministically greedy agent always picks the action with the largest \(\Psi\) value. However, when multiple actions have the same value, the latter agent samples an action uniformly from the set of best actions that involves the \textit{last} checker, which is the one fallen behind others. For both types of agents, if all possible moves are backward moves (i.e. negative forward distance, which is extremely unlikely but possible), the action with the least absolute heuristic value is chosen.

Through plays against humans, it was found that the performance of both greedy agents is robust to vastly different game states, because it always recursively searches for the longest possible move (which are often indiscernible by humans) without concerning any local strategies played by the opponent or even itself. However, for this very reason, greedy agents can hardly defeat any experienced human player in a game due to its lack of strategies, since human players can proactively plan and use strategies to ensure no checkers fall behind by sacrificing locally optimal plays or even make backward moves for assisting fallen-behind checkers. Figure 5 illustrates the critical weakness of heuristic-guided agents: with the game state on the left, Player 2 (red) will win in at most 5 moves; a heuristic-guided agent (blue) in this case would pick a move similar to the move noted with the black arrow and hence lose the game; however, an optimal play involves moving the checker \textit{backward} and use the other checker as a bridge (depicted by red arrows) and therefore win the game in 4 moves.

On the other hand, through a large number of matches between the two types of greedy agents,  it was found that the deterministic agent performs slightly better: after \linebreak playing 20,000 matches between the agents, where each agent took turn to start, the deterministic agent wins 9,908 matches while the stochastic agent wins 9,766 matches, with 326 games being draw due to move repetitions or moves that blocked the game progression. In addition, to reduce the likelihood of repeating or similar move sequences due to the nature of greedy agents, the first three moves for each agent are uniformly randomly selected. Therefore, the deterministic heuristic was used throughout our experimentation (described in later sections).

\begin{figure}[t!]
    \centering
    \includegraphics[width=0.35\textwidth]{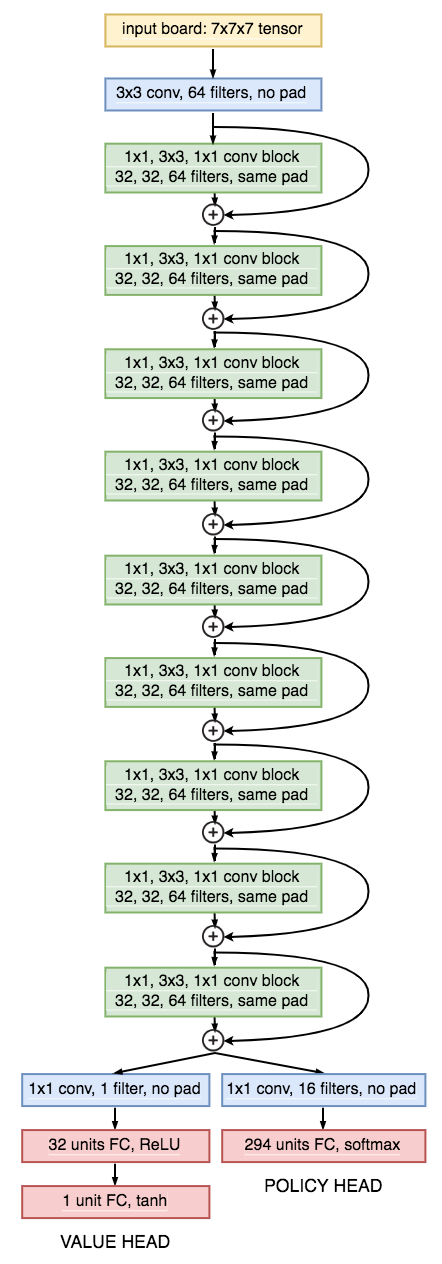}
    \caption{Network Architecture. All conv and FC layers use ReLU activation unless otherwise specified. All conv layers have a stride of 1.}
    \label{fig:fig6}
\end{figure}

\subsection{Network Architecture and Configurations}
The agent is represented as a single neural network which takes as input the current game state and outputs both the game-play policy (i.e. what actions should be taken with what likelihood) and the game state value (i.e. how good is the situation) for the current state from the perspective of the current player. The final neural network architecture is depicted in Figure 6. This residual convolutional network design was inspired by \cite{ref2} and \cite{ref27} for their great performance. Blue blocks in the figure represent single convolutional layers \citep{lenet} with ReLU activation \citep{alexnet} and a stride of 1. There are in total 9 convolutional blocks (represented as Green blocks in the figure), where each block contains a stack of 3 convolutional layers with \(1\times1\), \(3\times3\), \(1\times1\) kernels, 32, 32, 64 filters, and ReLU activations respectively. The skip connections (curved paths and add gates in Figure 6) between the convolutional blocks are adapted from ResNet \citep{ref27}. Batch normalisation \citep{ref26} layers are also added after each convolutional layer before the ReLU activation; no batch norm layers are added between fully connected layers. Also, bias was used on fully connected layers but not convolutional layers due to the bias introduced by batch norm layers. All weights in the network are initialised using the Xavier initialisation method \citep{ref28}. 

Convolutional layers with \(1\times1\) sized kernels \citep{ref27} were extensively used to reduce the number of parameters in the network, and they effectively perform dimensionality reduction on the feature maps while encouraging them to learn more robust features. In fact, compared to the original architecture described by \cite{ref2} from which our design is inspired, the current architecture has approximately 247,386 parameters, which is only around 22\% as much parameters. While the architecture design shown by \cite{ref2} has been proven effective, such networks are extremely hard to train given the amount of game-play data that can be generated without the use of massive amount of computing resources. As a result, excessive parameters will lead to significant overfitting in our experimental setting.

\paragraph{Network Input} The input to the network is a \(7\times7\times7\) tensor. The first six slices of the last dimension are six \(7\times7\) ID matrices (as illustrated in Figure 3) for the past three game states, where each game state is represented using two ID matrices, one from the perspective of the current player and the other from the perspective of the opponent, stacked in this exact order. The game states are then concatenated in sorted order where the most recent state is on the top. The last \(7\times7\) slice is a binary feature map full of value 0 if the current player is Player 1 and value 1 otherwise. When the game history contains less than 3 game states, the trailing slices of the tensor (except from the last one) will be filled with 0.

\paragraph{Network Output} There are two output-heads of the network, the policy head and the value head, following the practices used by \cite{ref2}. The value head takes the feature maps output from the residual blocks as input and outputs 1 real number in range \([-1, 1]\), which represents the evaluated game outcome from the perspective of the current player; a value closer to 1 means by playing from the current state, the current player is more likely to win than lose and vice versa. The first 49 elements in the output vector represent the prior probability for checker with ID 1 to move each possible position (49 of them in total) on the board; similarly, the second 49 elements represent the prior for the checker with ID 2 and so on. The conversion between vector indices to checker ID and position on board follows the unpacking order of checker ID, checker row, and checker column. For example, the \(181^{th}\) element in the vector refers to the probability of moving the checker with ID 4 to the 5th row and 6th column in the game matrix, where the ID of the checkers start from 1.

\subsection{Monte Carlo Tree Search}
In our work, MCTS is primarily used in two scenarios: to calibrate the action policy of the agent at training time, and to enhance the performance of the agent at test time. We have adapted the general MCTS framework from \cite{ref18} and \cite{ref19} with a several modifications inspired by \cite{ref2}:
\begin{enumerate}
    \item Rather than performing a Monte Carlo rollout at leaf positions of the tree (such as seen in the work of \cite{ref23} and \cite{ref24}) we instead use the neural network as a game state evaluator. This significantly speeds up the MCTS process by reducing the depth of the tree search as no playout till the end state is needed. For Chinese Checkers in particular, this is required as the depth of the game tree is unbounded.
    \item A variant of the Upper Confidence Bound (UCB) on Trees (or UCT) algorithm \citep{ref10} is used where the prior probability of an action is also taken into account in the exploration term of the search, which is slightly modified such that the search is more sensitive to the number of searches on a particular action and would therefore encourage more exploration on actions that are searched less often.
\end{enumerate}

\paragraph{Notation and Structure} Concretely, the Monte Carlo tree search used in this work involves four stages, which we referred to as Selection, Expansion, Backup, and Decision. To represent a game as a tree, we first store the raw game states using game matrices at tree nodes along with the current player and the prior probabilities of game state transitions, which can be predicted by the network in the Expansion stage. We then store the possible state transitions from the current state as edges directed away from the tree node. Each edge stores the starting and ending position of the move in the game matrix, the player who will make this move, and four important tree search statistics \citep{ref2}: \(N\), \(W\), \(Q\), and \(P\), where \(N\) is the number of times this edge has been visited during the search, \(W\) is the accumulated value (from the Backup stage) during the search, \(Q\) is the mean value (therefore \(Q = \frac{W}{N}\)), and \(P\) is the prior probability of selecting this move. We denote the \(N\), \(W\), \(Q\), \(P\) values of some edge \(k\) to be \(N_k\), \(W_k\), \(Q_k\), \(P_k\) respectively. The overall procedure of MCTS is to first perform a given number of \textit{simulations}, which are iterations of Selection, Expansion, and Backup stages; then, a final decision is made based on the edges that is directly incident to the root node, which represents the current game state.

\paragraph{Selection} The first step in a simulation is to iteratively select the locally optimal moves from the current game state (root of the search tree) until reaching a leaf node of the tree. However, note that as the search progresses down the tree, the simulated player is switched at every tree level. At any given state, the next action is determined by picking the outgoing edge with the maximum \(Q + U\) value, where 
\begin{equation} 
U_j = cP_j\frac{\sqrt{\sum_{k}^{}N_k}}{N_j +1} 
\end{equation}
for the edge \(j\), and \(c\) is a global constant controlling the level of exploration; through empirical analysis, the value of \(c = 3.5\) is used in our work. The value \(U\) serves to introduce an \textit{upper confidence bound} \citep{ref10} that the given action will be optimal. Intuitively, if an edge is rarely visited, then its \(U\) value will be bigger than that of the other edges, hence increasing the overall \(Q + U\) value and therefore encourages exploration. However, as the visit counts \(N\) increases, the search will still asymptotically prefer the edges with higher \(Q\) values as the significance of \(U\) decreases. 

While a commonly used expression for \(U\) on an edge \(j\) is
\begin{equation} 
 U_j = c\sqrt{\frac{2\ln{\sum_{k}^{} N_k}}{N_j}} 
\end{equation}
(also known as the UCT algorithm \citep{ref19}), it is unsuitable in the case of Chinese Checkers since it does not take into account the prior probability for selecting the edge \(j\). While this does not affect games with shallow or depth-bounded game-trees, in Chinese Checkers this may lead the agent to explore obviously unpromising moves (such as consecutive hops back) instead of better moves when the number of simulations is restricted. With the move prior \(P\), the search is more focused on more promising moves from the agent’s game-play experience. However, by introducing \(P\) in \(U\), the level of exploration of the search is constrained and game trajectories may depend undesirably heavily on past experiences. To mitigate this issue, the square root term in \(U_j\) is slightly modified such that the value of \(U_j\) is now more sensitive to \(N_j\).

\paragraph{Expansion} After iterative selection and arriving at a leaf node, the game state is evaluated by the neural network, which outputs a 294-dimensional vector \(\textbf{P}\) representing the prior for selecting the next moves from the leaf state, and a value \(\textbf{V}\) representing the evaluated game outcome from the perspective of the current player. The output vector is sparse, as the number of valid moves, denoted by \(n\), is limited. Then, \(n\) corresponding valid next states are created as nodes and are assigned to the current leaf node as children, while the \(P\) statistic of the new edges directed from the leaf node to the children will be assigned the corresponding move prior from vector \(\textbf{P}\). After expansion, the values of \(N\), \(W\), \(Q\) for the new edges are initialised to 0. This stage differs from a traditional MCTS procedure in that no playouts (aka ``rollouts'', ``unrolling'') is required, making the tree search more efficient.

\paragraph{Backup} Once the tree is expanded at the leaf node, the value \textbf{V} is back propagated along the path (i.e. the set of edges) from the root while updating the edge statistics of each edge along this path. For each edge \(e\) in the path:
\begin{enumerate}
    \item If the player of \(e\) (represented as an action) is the same as the player at the leaf node, the value \(W_e\) is added by \textbf{V}; Otherwise, \(W_e\) is subtracted by \textbf{V}. One exception, however, is that if the leaf state is a win state, then the direction of adding/subtracting \textbf{V} is inverted because the player at the leaf node has lost the game.
    \item The value \(N_e\) is incremented by 1.
    \item The value \(Q_e\) is updated to be \(\frac{W_e}{N_e}\).
\end{enumerate}
Figure 7 summarises the above three stages involved in a simulation.

\paragraph{Decision} After the search is finished with many iterations of Selection, Expansion, and Backup, the agent randomly samples a move from the root state using the exponentiation of the visit counts as prior probabilities; if we denote the post-search prior to be \(\textbf{P*}\), then for an edge \(e\), its probability of being selected is:
\begin{equation} 
\textbf{P*}_e = \frac{N_e^{\frac{1}{t}}}{\sum_{k}^{}N_k^{\frac{1}{t}}}
\end{equation}
where \(t\) (adapted from \cite{ref2}) is a temperature parameter controlling the level of exploration. Intuitively, a value \(t > 1\) will lead to a smoother probability distribution over the edges, while a \(t\) close to 0 will lead to a deterministic choice, since \(\max_{e}(p_e)\) will be close to 1. In our work, \(t\) is set to 0 during evaluation and testing for achieving deterministic plays and is set to a value bigger than or equal to 1 for the initial moves of the self-play games for encouraging diversity in the generated training data. After performing the tree search for a particular state, the entire search tree is discarded and a new search tree will be built for each new game state. More details are presented in Experiments and Results. 

\begin{figure*}
    \centering
    \includegraphics[width=0.8\textwidth]{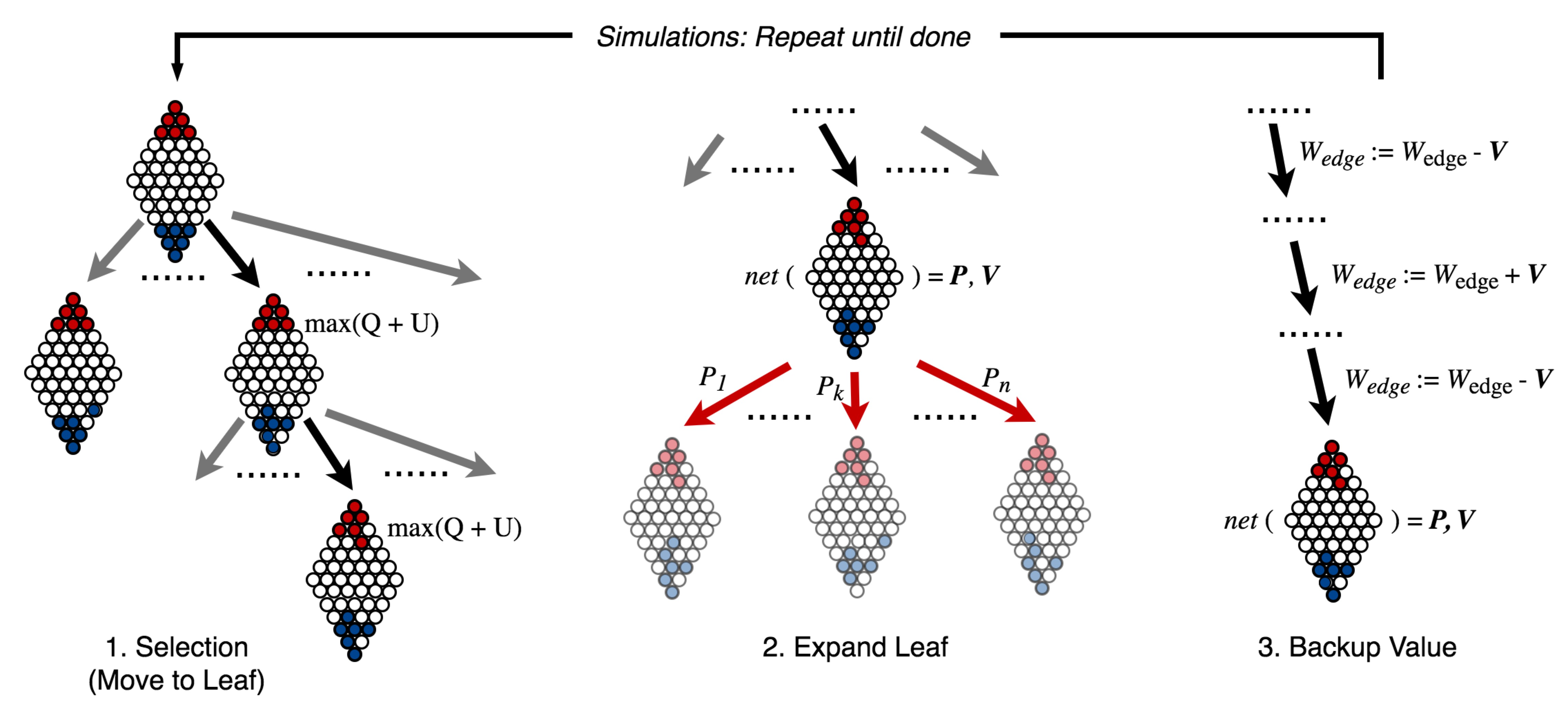}
    \caption{Summary of the simulation process of the Monte Carlo Tree Search used in this work.}
    \label{fig:fig15}
    % refered as figure 7
\end{figure*}

\subsection{Data Generation and Training Pipeline}
In summary, the agent is trained using a two-stage training pipeline: it is first trained with histories of self-plays guided by the deterministic heuristic, and is then further optimised through reinforcement learning using \linebreak MCTS-guided self-plays. It is worth mentioning that for both stages of the pipeline, no human game plays are used, and the training data are generated entirely through agent self-plays.

\paragraph{Labels and data augmentation} Each training example \linebreak comprises the \(7\times7\times7\) tensor representing the current game state as input, and the expected game state value and post-search policy vector as the ground truth. The game state value label, denoted as \textbf{V*}, will always be either -1 or 1 representing a loss or a win from the perspective of Player 1. There is no draw in the game, as such games are only possible due to meaningless repetitions or irrational moves. The policy ground truth, denoted by \textbf{P*} will be a 294-dimensional vector representing move prior. During our experiments, the size of the training set is augmented to be twice as large by flipping each slice of the input tensor along the last dimension around the bottom-left-top-right diagonal (i.e. mirroring the board).

\paragraph{Optimisation Objective} The loss function for the network is an unweighted sum of the losses at the two output heads of the neural network. With each training example, the loss for the value head is calculated using mean squared error loss between predicted game state value \textbf{V} and ground truth \textbf{V*}, while the loss for the policy head is calculated using cross entropy loss between predicted move prior \textbf{P} and policy ground truth \textbf{P*}. L2-regularisation is also added to the loss term. In summary, loss \(L\) is given by:
\begin{equation} 
L = \textbf{P}^T log(\textbf{P*})+\|\textbf{V}-\textbf{V*}\|_2 + \lambda\|\theta\|_2
\end{equation}
where \(\theta\) refers to the parameters of the network and \(\lambda\) is a constant controlling the strength of L2 regularisation. 

\paragraph{Training on heuristic-guided self-plays} By directing the greedy agent to play against itself without any form of look-ahead search and game state evaluation, we are able to quickly generate a substantial amount of training data within a short amount of time. On average, it takes approximately 2 minutes to generate 5,000 self-play games when fully utilising two Intel i5 2.7 GHz CPU cores. 

In this setting, each game state is a training example where the state value ground truth is the final outcome of the game (with the value negated when switching players); the policy ground truth is calculated by assigning the reciprocal of number of moves with most forward distance to the corresponding optimal move indices in a 294-dimensional, zero-filled vector. Heuristic-guided training is done in a supervised manner where self-play data are first generated and the agent is then trained on them over a fixed number of epochs. To increase the diversity of the data and reduce memory consumption, our implementation in fact generates the training set on-the-fly with approximately 15,000 new games per iteration and uses a smaller number of epochs per iteration (5 per iteration). However, there are two important factors that makes the raw heuristic-guided self-play data less useful. First, even if each move in the games is sampled rather than picked deterministically, one can still observe frequently occurring game-play patterns due to the agent’s locally optimal strategy. This may lead to undesirably high correlations between training examples. To mitigate this, we introduced three measures to add stochasticity to the data generation process:
\begin{itemize}
    \item Randomising the starting positions of all checkers (which gives rise to more than \(9\times10^{10}\) possible starting states).
    \item Forcing the agent to take random initial moves which causes subsequent moves to be substantially different.
    \item Only retaining a very small portion of the generated training examples by random sampling. The percentage of data kept is around 3-5\% depending on need. 
\end{itemize} 
The ratio between randomising starting checker positions, randomising starting moves, and normal plays is set to be 5:3:2, but other variations are also possible. Secondly, the heuristic-guided agent only focuses on locally optimal moves, which means the agent can easily make forward-only moves and avoid some globally optimal plays that require horizontal or backward moves as depicted in Figure 5. To mitigate this,  the first stage of the training pipeline is stopped before the model is trained till convergence to increase the malleability of the network for further training through reinforcement learning, in which stage the training data is harder to generate. Another minor drawback of heuristic-guided self-plays is that there is a very small probability that a games does not finish due to blocking checkers. Since draw games are undesired in the training set and in general, an early stopping mechanism is introduced where the maximum length of each self-play game is limited of 0.1s, and draw games are discarded.

\begin{figure*}[t]
    \centering
    \includegraphics[width=0.8\textwidth]{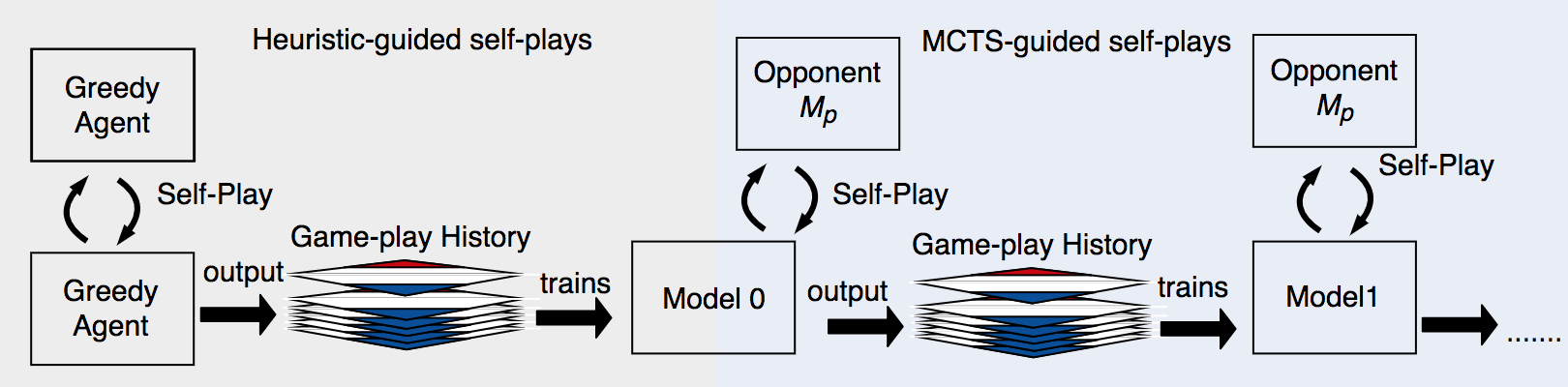}
    \caption{Summary of our training framework.}
    \label{fig:fig7}
    % referred as figure 8
\end{figure*}

\paragraph{Training on MCTS-guided self-plays} Once the network is properly initialised through supervised learning, it is further trained through self-play reinforcement learning, where MCTS is used at each state to generate an improved move prior \textbf{P*} as the policy label. Our reinforcement process is described as follows. Starting from the initial version, we maintain the current model \(M\), the best model \(M^*\), an opponent \(M_p\), where \(M^*\) and \(M_p\) are initialised to be \(M\). Starting from the first episode and for each episode, we match \(M\) against \(M_p\) to generate a fixed number of self-play games, and then randomly sample 50\% of the generated data to break correlation and use them to train \(M\). After a number of epochs of training, the newer version of \(M\), is evaluated against \(M^*\) and a human player on a certain number of games. If \(M\) defeats \(M^*\) by more than 55\%, then \(M^*\) and \(M_p\) can optionally be updated with \(M\), depending on whether \(M\) is overfitting to a policy, which is measured by its performance increase against a versatile and dynamic human player, or whether \(M_p\) is chosen to be the greedy player, in which case \(M_p\) will not be updated. Note that variations to the above reinforcement process is possible. For example, it is possible to have \(M^*\) and \(M_p\) to be the same model, to have a different win rate threshold, to have \(M^*\) instead of \(M\) match against \(M_p\) to generate training data, or to discard the use of \(M_p\) and \(M^*\) altogether so that the training is always iterating on the only model \(M\). While initial experiments show that these variations bring little difference in terms of a trained model’s win rate against human players, whether differences in the reinforcement process can profoundly impact the training and performance of the model remains an open question worth further exploration. 

In contrast to the generation of heuristic-guided self-plays, the rate at which MCTS-guided plays are generated are remarkably slow. On average, it takes around 120 minutes to generate 180 self-play games with 175 simulations per Monte Carlo tree search on 12 CPUs @ 3.7GHz after the generation procedure has been parallelised, yielding an average of 1.5 minutes per game. Another pitfall was that MCTS-guided plays also suffered the lack of game-play diversity. To mitigate this, the level of exploration by the agent during training is increased using the following strategies:
\begin{itemize}
    \item First 6 moves of the game are randomly chosen
    \item Subsequent 10 moves use a large temperature \(t\), typically 1 or 2, for  post-search decision
    \item The remainder of the game is played deterministically with \(t\) = 0.01, so that the best possible moves are always chosen by the agents.
\end{itemize}

The number of moves to play randomly and exploratively are determined empirically such that the post-move game state are likely to occur in a normal human-to-human match while significantly contributing to the diversity of future game states. In addition, before performing the very first simulation of MCTS (i.e. when the search tree is empty), a small noise vector with the same dimension as the number of valid actions is drawn from the Dirichlet distribution where \(\alpha = 0.03\), and the vector is added to the move prior at the root node with weight 0.25, following the practices from \cite{ref2}. This encourages all immediate next moves to be tried. Figure 8 summarises the overall training framework.

\subsection{Alternative Approaches}
In addition to the approach described above, we also experimented two alternative approaches for comparison: Deep Q-Learning and \(tabula rasa\) reinforcement learning.

\paragraph{Q-Learning} Like all traditional reinforcement learning \linebreak approach, Q-learning consists of five major components: a set of all possible states, a set of all possible actions, a transition probability distribution describing the probability over what state it would transit to by taking a specific action in a given state, a discount factor controlling the balance between gaining short-term reward and long-term reward and a reward function which takes a state and an action and outputs the corresponding reward of taking this action in this state. By making a move in a given state, the agent will get a positive or negative immediate reward and the ultimate goal of Q-Learning is to maximize the accumulated reward. In other words, we are trying to maximize the value of:
\begin{equation} 
R(S_0,a_0) + \delta R(S_1,a_1) + \delta^2 R(S_2, a_2) + ...
\end{equation}
where \(R\) is the reward function mapping a particular state \(S_i\) and the action \(a_i\) to a real-valued reward, and \(\delta\) is a discount factor with range \(0 \leq \delta \leq 1\). To manage this accumulated reward with more ease, the above formula could be transformed into a recursion by defining \(Q(S, a)\) as the accumulated reward in state S when we perform action \(a\). Thus, the action we choose in a certain state would be the action resulting into the maximum \(Q\) value:
\begin{equation} 
Q(S, a) = R(S, a) + \delta \sum{P_{sa}(S') \max_{a} Q(S', a')}
\end{equation}
where \(S'\), \(a'\) is the next state and action, \(P_{sa}\) gives the state transition probability with a given state, and \(\sum_{S}P_{sa}(S) = 1\) and \(P_{sa}(S) \geq 1\).
To calculate the actual value for all \(Q(S, a)\), a method called Value Iteration is developed. By initializing all \(Q(S, a)\) with arbitrary value, we consistently updates all \(Q(S, a)\) with the difference of current \(Q(S, a)\) and the new \(Q(S, a)\) we estimated through actually performing action a in state \(S\) and observing the immediate reward. The details for this algorithm is provided in Algorithm 1.
\begin{algorithm} \caption{Value Iteration}
\begin{algorithmic}[1]
\STATE Initialize \(Q(s, a)\) arbitrarily 
\REPEAT 
\FORALL{episode} 
\STATE Choose \(a\) from \(s\) using policy derived from \(Q\)
\STATE Take action \(a\), observe \(r, s\prime\)
\STATE \( Q(s, a) \gets Q(s, a) + \alpha[r + \gamma \max_{a \prime} Q(s\prime, a\prime) - Q(s, a) ] \)
\STATE \( s \gets s\prime \)
\ENDFOR
\UNTIL \(s\) is terminal
\end{algorithmic}
\end{algorithm} \\

Although mathematically there is a guarantee that \linebreak Value Iteration would converge to the actual Q-value, in so many cases the number of all possible state and action pair is incredibly large, which could not be fitted into the memory. To solve this problem, we adapted the approach from \cite{ref35}, which uses a convolutional network that takes the state as the input and predict the \(Q(s, a)\) values.

The major difference between our game and Atari \linebreak games is that Atari games has a deterministic environment, which means given a certain state and its history we can actually predict what state the environment will transit into. However, in our case the state transition does not only depend on the current state and the action the agent picks but also the action the opponent picks. In this scenario, simply randomizing the opponent’s move is not particularly preferable, since random move normally leads to tie and does not provides very useful guidance for the agent. In order to help the agent to learn the optimal policy efficiently, we need to set up an opponent with good performance. Thus the greedy policy agent is used as its opponent.

With this pre-defined opponent, we can set up the transition probability and the rewarding mechanism. For simplicity, the transition probability given a state and the chosen action is one divided by the number of valid moves so that it promotes generality. To encourage the agent to move forward and learn to move backward for long-term benefit, we designed the rewarding system to give a positive reward equivalent to the number of rows it jumps across in the forward direction in this move and give a negative reward equivalent to the number of rows it jumps across in the backward direction, multiplied by 0.01. And when all the checker pieces arrived the other side of the board resulting into victory, we give the agent a very large reward which in our case set to be 10. With such rewarding mechanism, it is shown that the agent has learned the technique of moving a small step backward so that it could gain larger rewards in the next few moves.

As for the network architecture, the major structure is the same as the structure we mentioned in the previous subsection. To fit into Q-Learning usage, the input and output layer are modified. Since one training example only includes the Q-value for one state-action pair and the network outputs the Q-values for all actions associated with the input state, we need to add a mask in the final output layer to make sure the loss will only include the action we choose. This is accomplished by setting up an additional input layer with the same dimension as the output layer as the mask. During the updating process, only the index representing the action we choose is set to be 1 and all the other elements in the mask input vector are filled with 0. In this way, regression can be performed for a single output neuron. To make a prediction, we use a vector all filled with 1 as the mask input vector. 

To generate the training data, we need the agents to actually play the game and get the reward from the environment. Thus, a pair of the state and action with their corresponding Q-value would be one training example. There are generally two ways to choose action for the agent during the process of generating training data. The first way is randomly choosing one action from the valid action set, which turns out to be a good option when the training process just started as it emphasizes on exploration. The second way is picking the action with the highest Q-value from the network’s output, which will be effective during the later stage of training as it concentrates on exploitation. To combine the advantages of these two approaches, we used a method called epsilon-greedy policy, where we keep a probability \(\epsilon\) for choosing an action randomly instead of deterministically based on the network's output. Initially, \(epsilon\) is set to be 1 and as the training process going we gradually decrease the epsilon until the epsilon reaches 0.1 so that in the later stage of the training process we will have a much higher probability to choose the action based on the network's output and at the beginning we are more likely to choose the action randomly. A graphical illustration of the epsilon schedule is illustrated in Figure 9.

\begin{figure}[t]
    \centering
    \includegraphics[width=0.47\textwidth]{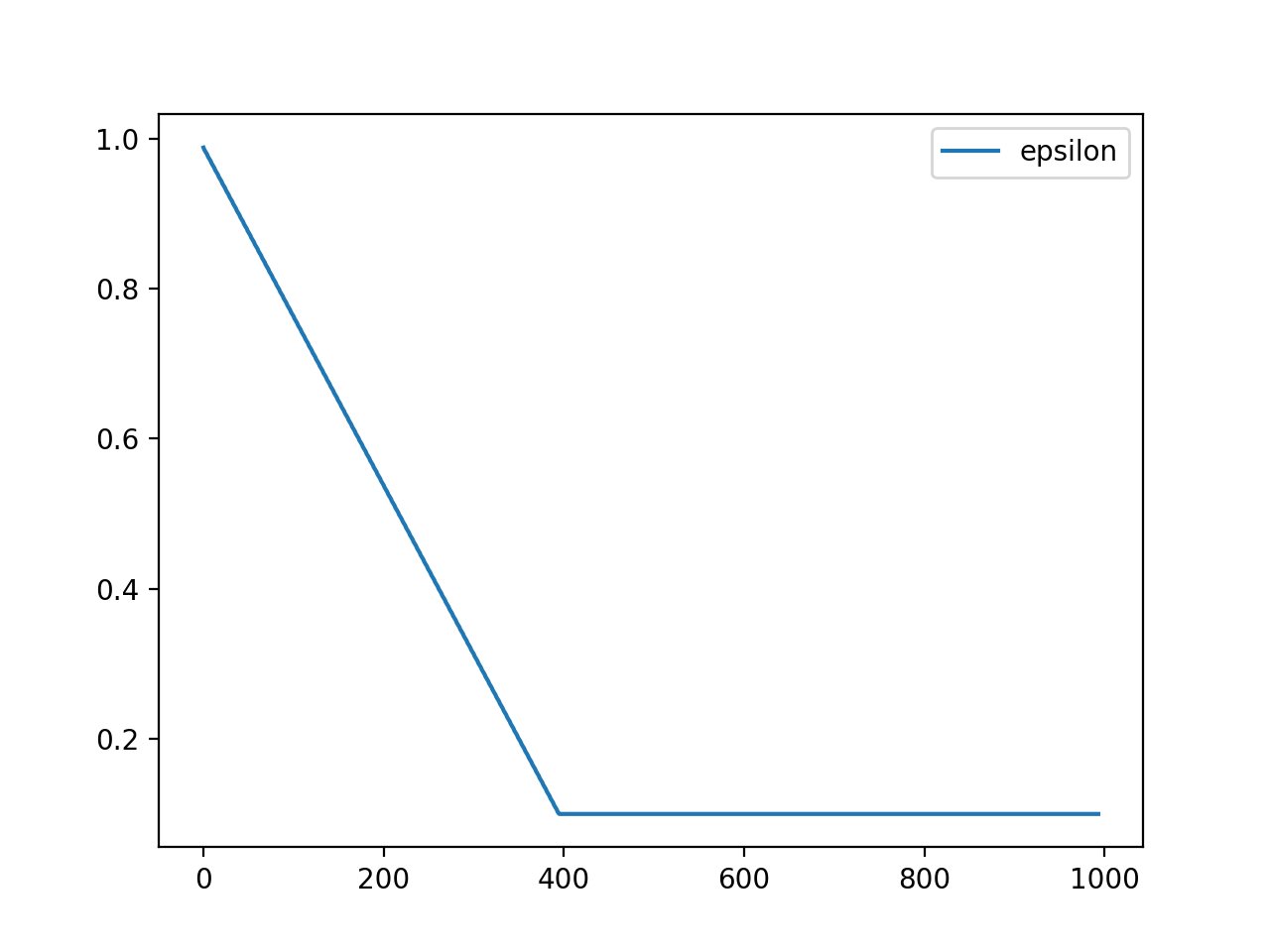}
    \caption{Q-Learning Epsilon-Greedy schedule over training. \(x\)-axis represents the number of episodes divided by 10,000. \(y\)-axis represents the value of \(\epsilon\).}
    \label{fig:fig12}
    % referred as figure 9
\end{figure}

In order to perform regression, we need to define a loss function, and the commonly used Mean Square Error is problematic in this case because larger errors are emphasized over smaller errors due to the square operation; large errors will lead to radical change of the network which in turn change the target value dramatically. In comparison, Mean Absolute Error addresses this problem, but the penalty for larger errors is insufficient. To keep a balance between these two loss functions, we used a special loss function called Huber Loss, which uses Mean Square Error for low error and Mean Absolute Error for large error. The equation of Huber Loss function is provided below:
\begin{numcases}{ Huber(a)=}
  \frac{1}{2} a^2, & for $|a| \leq 1$\\
  |a| - \frac{1}{2}, & otherwise
\end{numcases}

\begin{table*}[t]
    \centering
\begin{tabular}{ l l }
  \hline
  \multicolumn{2}{c}{Final Training Configurations} \\
  \hline
  Input dimension & \(\rm I\!R^{7 \times 7 \times 7}\) \\
  Output dimension policy vector & \(\rm I\!R^{294}\) \\
  Output dimension policy vector & \(\rm I\!R\) \\
  Dirichlet noise weight & 0.25 \\
  Dirichlet parameter \(\alpha\) & 0.03 (as vector) \\
  Game move limit & 100 \\
  Reward policy & 1 for win, -1 for loss \\
  Batch Size & 32 \\
  L2 regularisation \(\lambda\) in heuristic stage & 1e-4\\
  L2 regularisation \(\lambda\) in MCTS stage & 5e-3 \\
  Learning rate & 1e-4 \\
  Optimizer & SGD + Nesterov \\
  Number of epochs in heuristic stage & 100 in total \\
  Number of epochs in MCTS stage & 5 per episode \\
  Number of self-play games in heuristic stage & 15000 in total \\
  Number of self-play games in MCTS stage & 180 per episode \\
  Number of evaluation games & 24 \\
  Number of Monte Carlo simulations & 175 \\
  Temperature parameter \(\tau\) for exploratory play & 2 \\
  Temperature parameter \(\tau\) for deterministic play & 1e-2 \\
  Number of moves in exploratory play & 5 per player \\
  Tree search exploration constant \(c\) & 3.5 \\
  Initial random moves & 3 per player \\
  Hardware CPU & 12 Intel i7 \\
  Hardware GPU & 1 Nvidia GTX 1080 \\
  Memory limit & 32 GB \\
  \hline
\end{tabular}
    \caption{Final training configurations for the main approach.}
    \label{tab:my_label}
\end{table*}

\paragraph{Tabula rasa Reinforcement Learning} We experimented \linebreak with the training approach similar to that described in \cite{ref33}, where an agent, denoted by \(M_0\), is trained based solely on MCTS-guided reinforcement learning. Instead of using a guidance, \(M_0\) is trained directly from the second stage of the current training pipeline, starting from scratch and total random plays and uses MCTS to guide its plays against itself to generate training data, where the policy labels and value labels are the post-tree-search move prior and the actual game outcome respectively. All other training configurations are adapted from Table 1, with minor necessary modifications such as learning rate and regularisation strength.

The single most important problem with \textit{tabula rasa} learning in Chinese Checkers is that when agents are \linebreak trained starting from random play, the self-play game may never finish. To address this problem, we introduced two mechanisms: a soft limit on the total number of moves, and a different reward policy that encourages forward movement without requiring the game to reach a win state. The soft move limit involves setting a constant \(T\) to be the initial allowed number of moves of the game; if no \(progress\), which is defined by the net increase in the number of checkers moved into the opponent’s base, has been made by either player within \(T\) moves, then the game is terminated. When progress has been made, the allowed number of moves would be incremented by \(T\), and the process repeats as the game continues. The outcome of the game is determined by the following reward policy: once the game terminates, both players’ total forward distance in the game is first computed to be the sum of the number of rows that a player’s checkers have advanced in the board, and then the absolute difference \(D_p\) between the distances is computed. The player with a larger forward distance is considered the winner if \(D_p\) is greater than a pre-set threshold \(D_t\), in which case this player is given a reward of 1 while the other player receives a reward of -1. When \(D_p < D_t\), the game is considered draw, which is not a possible outcome in the two-stage training framework. Overall, this reward policy introduces the distinction between “good” or “bad” moves, instead of having a reward of 0 everywhere. In our implementation, \(D_t\) was set to 3, and the soft limit \(T\) was set to 100.

\section{Experiments, Results, and Analysis}
In this section, we first present and discuss the training and the performance of the agent, and we present the details and results of the experiments conducted on several important aspects of the training framework including game initialisation techniques and other fine-grained hyper-parameters, and we discuss how each of these aspects may affect the agent’s performance. In addition, we compare the current training framework with other approaches including Q-Learning and tabula rasa reinforcement learning. For most experiments, we use the greedy agent as a baseline for comparison unless otherwise stated, since the heuristic is always robust and deterministic (though not necessarily optimal) under various game states and its local optimality means it has a constant performance.

\begin{figure}[t]
    \centering
    \includegraphics[width=0.45\textwidth]{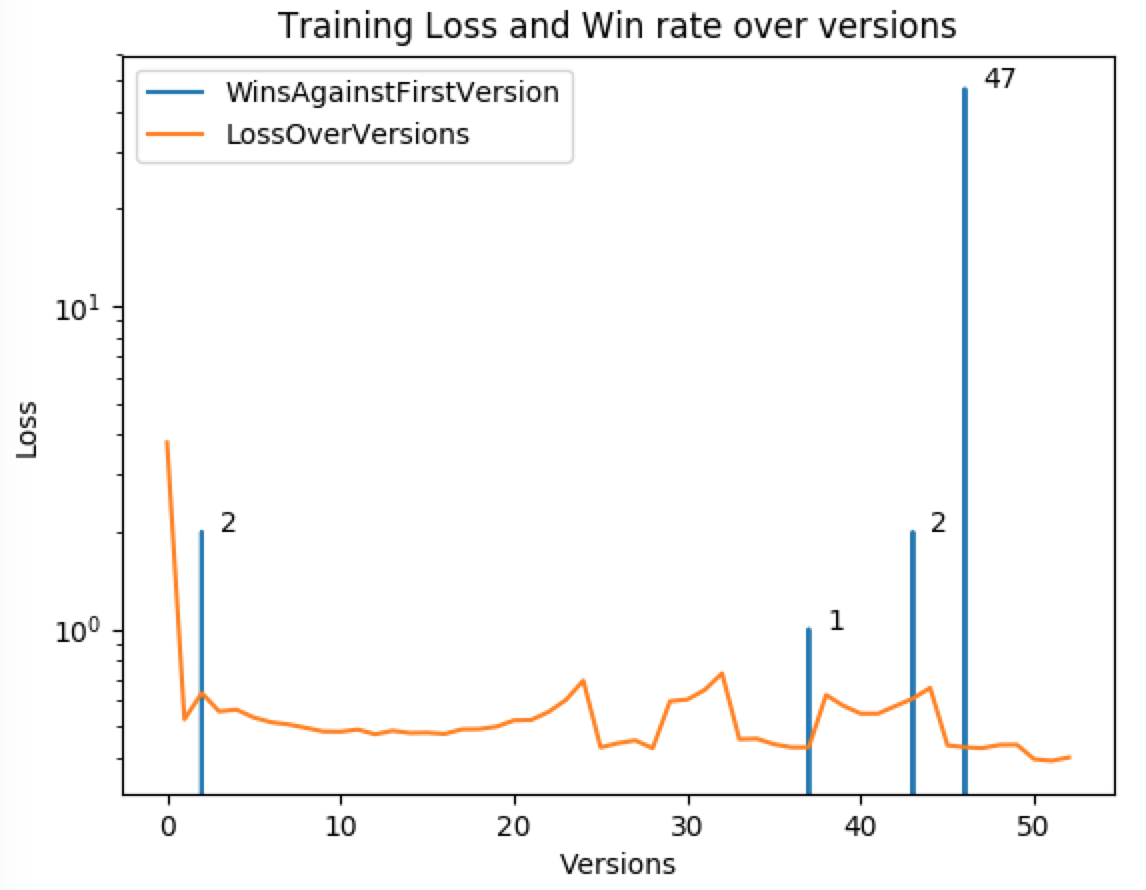}
    \caption{Agent overfitting to a specific policy, indicated by sudden increase in win rate against another model. Note that log-scale is used for y-axis due to the differences in data magnitudes and spikes in the model’s loss.}
    \label{fig:fig8}
    % referred as figure 10
\end{figure}

\subsection{Training}
\paragraph{Main approach} The final training configuration are summarised in Table 1. Training is done on 12 CPUs and 1 NVIDIA GeForce 1080 GPU: for the first stage of the training pipeline, the agent is supervised by heuristics for 100 epochs; in each epoch, 15,000 heuristic-guided self-plays are generated, yielding approximately 40,000 to \linebreak 50,000 training examples given average game length of 40 moves, the two-fold data augmentation, and 4\% data retention rate. The agent is then trained on these examples for 2 iterations before continuing to the next epoch. For the second stage of the training pipeline, 180 self-play games are generated in each episode from self-play games using the current best model \(M^*\) with 175 Monte Carlo simulations for each action. By combining training data from the previous episode and randomly sampling 50\% of them, around 18,000 training examples are retained for each episode. During the second stage of training, however, the model can easily overfit to the generated examples due to the relatively small number of training examples compared to the first stage. Figure 10 depicts the loss curve of an overfitting model and its performance when matched against the initial version of itself. It can be observed that as the loss decreases, the agent in fact learns a counter-policy specifically for its opponent, which is the initial model out of the first stage of the training pipeline. The overfitting is indicated by the sudden increase in the number of wins at version 46, but very few wins by the earlier versions. In order to prevent overfitting, the agent is trained on the examples generated in each episode for only 5 epochs before continuing to the next episode. Figure 11, on the other hand, illustrates the model’s learning progression with regularisation proportionate to the number of generated examples; it can be observed that the number of games won against the first version is gradually increasing. Agent evaluation, as described in the previous section, is done after each episode where the model is matched against the best agent \(M^*\) (where no \(M_p\) is kept) so far on 24 games and we update \(M^*\) with the current agent if the newer agent is able to defeat \(M^*\) by at least 14 games. 
\begin{figure}[t]
    \centering
    \includegraphics[width=0.45\textwidth]{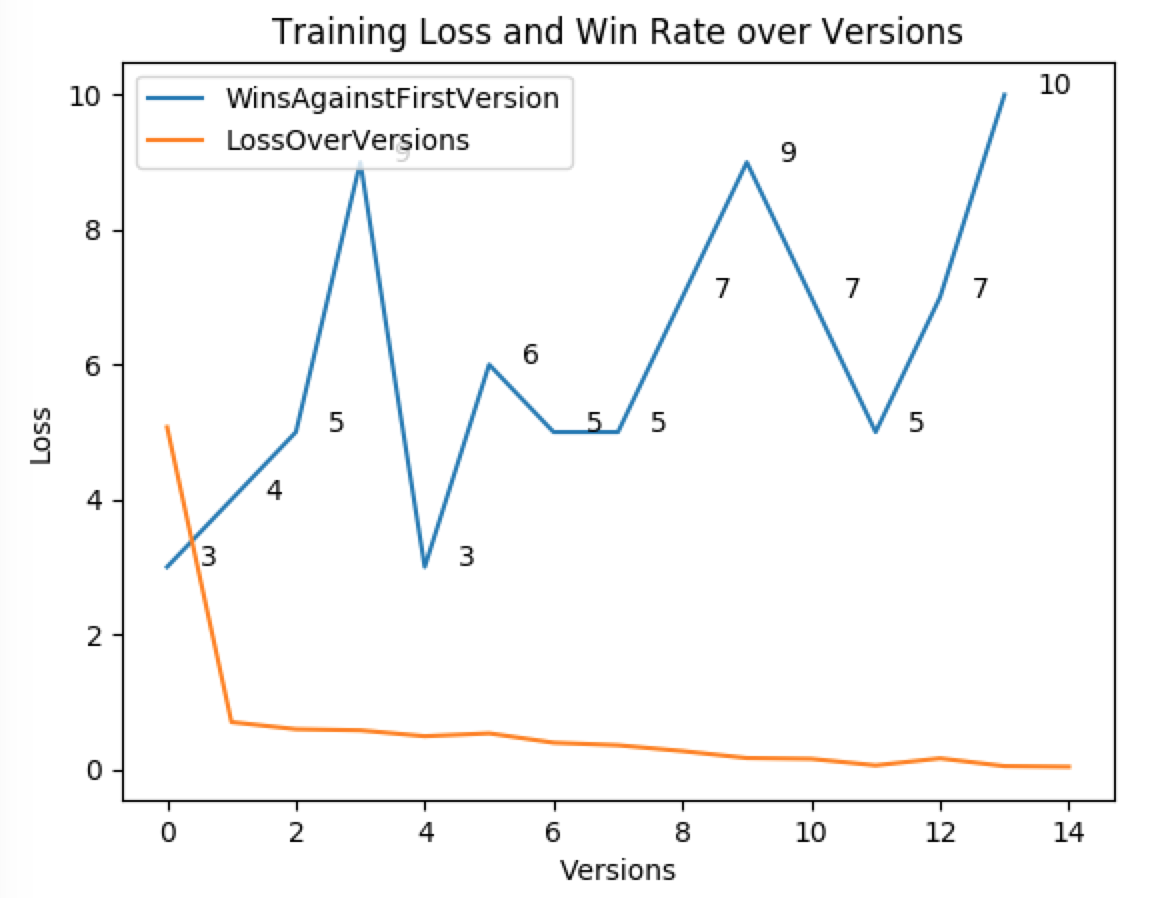}
    \caption{Loss curve and model performance of the current model architecture with suitable regularisation.}
    \label{fig:fig9}
      % referred as figure 11
\end{figure}

\paragraph{Q-Learning} During training data collection, the game is set to the initial state and the training agent is set to play against the greedy policy agent until the end of the game using the epsilon greedy policy mentioned in the previous section. For every move the training agent made, a tuple of \((S_t, a_t, R_t, S_{t+1})\) is collected, where \(t\) is the time step.

In order to train the model more efficiently, we also set a move limit, which means the total amount move made by the agent reaches this limit, this round of game will terminate. The reason behind this move limit is that randomly choosing move will not typically lead to the normal termination state which is either all checker pieces of the agent moves to its opponent’s base or other way around and we only want to keep those moves that may leads to improvement to the model. During our training process, we set the move limit to be 40 which is slight higher than the average number of moves greedy algorithm agent takes to finish the game against itself.

To prevent the model from divergence, we adapt the method called experience replay into our training process: in each iteration, instead of updating the network based on the move that were just made, we keep a finite training data pool (in our case the limit is set to 1,000,000 training examples) and in each iteration we randomly pick a mini-batch of training data (in our case the size of mini-batch is 32) from the training data pool to update the network. Before training begins, we pre-fill the training data pool with 5000 training examples, so that it is highly unlikely to include successive states in the mini-batch.

In summary, for each epochs of the training process, we initialise a game with the initial state and let the training agent play against the greedy policy agent to the end using epsilon greedy policy. Every time the agent makes an action, the \((S_t, a_t, R_t, S_{t+1})\) tuple is added into the training data pool. Once the size training data pool reaches a threshold, a mini-batch is drawn randomly from the pool to update the network. Throughout the entire training process, 10,000,000 epochs are run and around 400,000,000 training tuples are generated. The mean absolute error during training is illustrate in Figure 12.

\begin{figure}[t]
    \centering
    \includegraphics[width=0.45\textwidth]{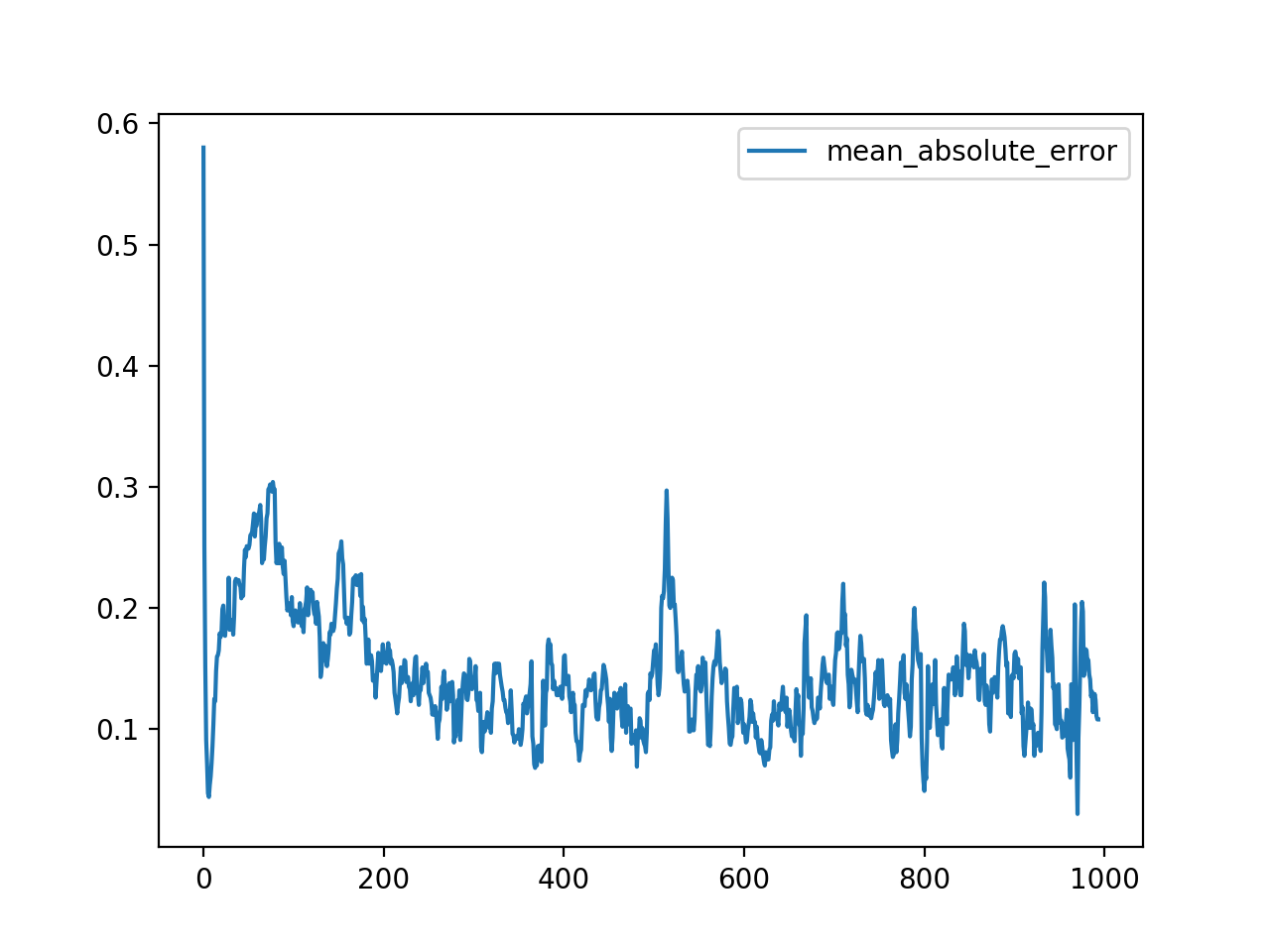}
    \caption{Q-Learning Mean Absolute Error for every 10,000 training episodes.}
    \label{fig:fig13}
    % referred to as figure 12
\end{figure}

\begin{figure}[t]
    \centering
    \includegraphics[width=0.25\textwidth]{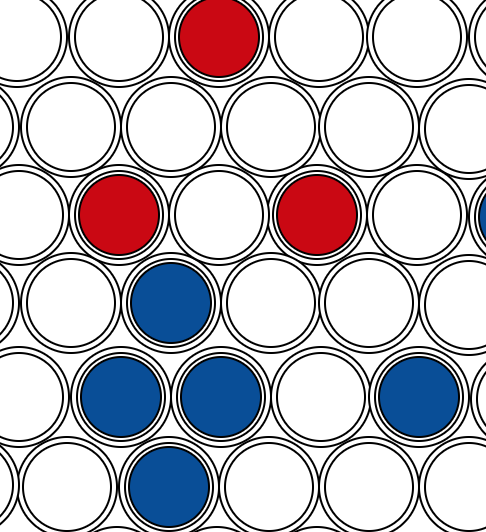}
    \caption{Typical huddle-formation (blue) to prevent the opponent (red)’s checkers from hopping.}
    \label{fig:fig10}
     % referred to as figure 13
\end{figure}

\paragraph{Tabula rasa Reiniforcement Learning} The training procedure for this approach is similar to that of the main approach, except that the first stage of the training pipeline is removed. Most training configurations are identical to that described in Table 1, except that the reward policy is different as described in the previous section. Starting from a random model \(M_0\), in each episode we generate 180 self-play games where each step uses 175 Monte Carlo simulations, yielding approximately 7,000 training examples per episode and they are iterated for 5 epochs before continuing to the next episode, identical to the stage 2 training pipeline of the main approach. An additional opponent model \(M_p\) is kept for generating training data, and is replaced with the newest version of M0 when the win rate exceeds the pre-defined threshold. 

\begin{table*}[t]
    \centering
\begin{tabular}{l *{5}{c} }
Opponent & Agent Wins & Agent Losses & Game Abandoned & Total & Agent Elo Rating Change \\
\hline
Greedy Agent &271 (90.3\%)& 28 (9.3\%)   & 1              & 300   & +392  \\
Human        &63 (63\%)   & 32(32\%)     & 5              & 100   & +111  \\
\end{tabular}
    \caption{Performance of the main approach agent.}
    \label{tab:my_label2}
\end{table*}

\subsection{Performance}
\paragraph{Main approach} The performance of the agent is summarised in Table 2. The computing resource used for testing is one Intel i7 CPU @ 3.7GHz. In each game, the thinking limit for each step is set to 180 Monte Carlo tree searches, leaving a thinking time of approximately 3 seconds. In total, the trained agent was tested by playing 300 games against the agent directly following the greedy heuristic and 100 games against a proficient Chinese checkers human player. The trained agent was able to achieve roughly 90\% win rate against the greedy player and 63\% against a human player. Abandoned games are infrequent and are mostly due to the agent refusing to move its last checkers out of its base to avoid its imminent loss, which can be foreseen from its lookahead search. Overall, it is reasonable to conclude that the agent is very robust against the greedy agent and strong against experienced human players, and we argue that this is partially due to the effective look-ahead Monte Carlo tree search, where the agent can deliberately interfere with the long, consecutive hops that the greedy agent is best at. For example, the agent would aim to block or break opponent’s checker ``bridges'' as they are forming, either by intruding with its own checkers or removing its checkers that are part of the bridges. When played against humans, the agent exhibits greedy-like traits such as seeking to take the longest hops, but it also learns to form strategies such as the bridge and to occasionally sacrifice locally optimal moves for a better long-term strategy. For example, in a local game state depicted by Figure 13, the agent (blue) would prefer to maintain its huddle-formation (where the 4 checkers stick together) to block the red checkers from hopping over, instead of taking the locally optimal move of advancing its checkers and breaking the cluster. 

However, the agent also has a few weaknesses. When exploration is disabled during testing (with temperature \(t\) set to 0.01 for Monte Carlo tree search and no initial random moves), we can hardly observe diversity in its starting strategies. In addition, it rarely takes several locally sub-optimal moves in a row for a better global strategy. We conjecture that the agent is partially bias towards greedy-moves due to its heuristic-guided initialisation; however, it is believed that this condition can be mitigated and its performance can be further improved through further self-play reinforcement learning.

\begin{figure}[t]
    \centering
    \includegraphics[width=0.47\textwidth]{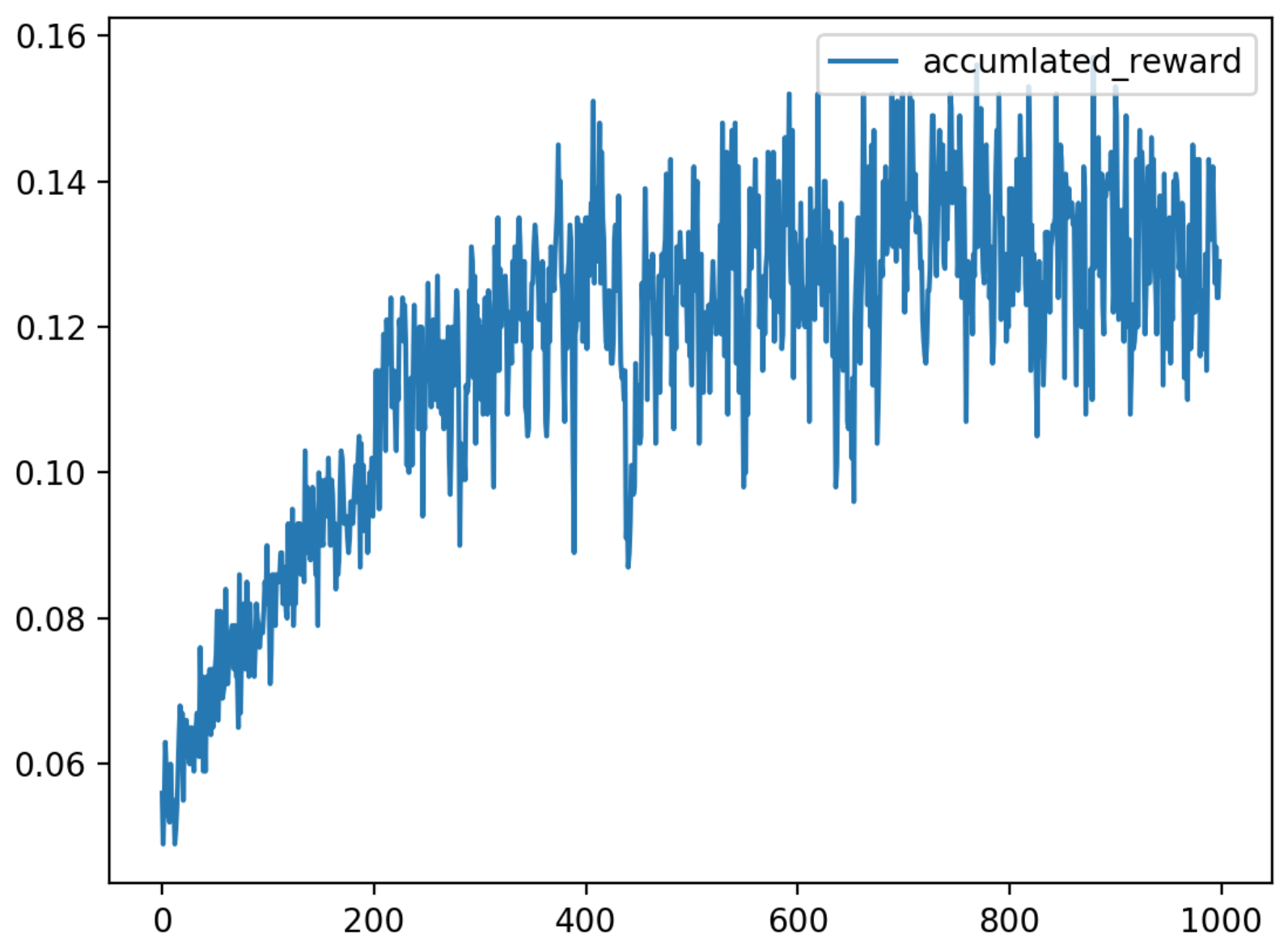}
    \caption{Q-Learning accumulated reward from testing after every 10,000 episodes of training. \(x\)-axis represents the test game number, and the \(y\)-axis represents the accumulated reward.}
    \label{fig:fig14}
\end{figure}

\paragraph{Q-Learning} After training to convergence, we evaluate the performance of Q-Learning agent by playing against the greedy policy agent. The accumulated reward after each game played at the end of every 10,000 episode (totalling 1,000 games) is illustrated in Figure 14, where the y-axis represents the reward value. By observing the game, we find that the Q-Learning agent successfully learns some basic strategies such as sacrificing checker progress by moving back checker pieces for better long-term plays. During the beginning of the game, it even outperformed greedy policy agent dramatically. However, as the game goes on, the performance of the agent drops significantly as it starts to focus on one checker only and later randomly acting. By analyzing the training data, we find that this is due to the fact that by following the epsilon-greedy policy, the agent tends to ignore some checker pieces to maximize the temporary reward. As greedy policy agent quickly blocks the path of the checker pieces left behind, the game almost always ends with a tie, which leads to insufficient training data for the later part of the game. We argue that the problem is mainly due to the nature of the game and the complexity of it such that Q-learning may not be suitable. 

\begin{figure}[t]
    \centering
    \includegraphics[width=0.4\textwidth]{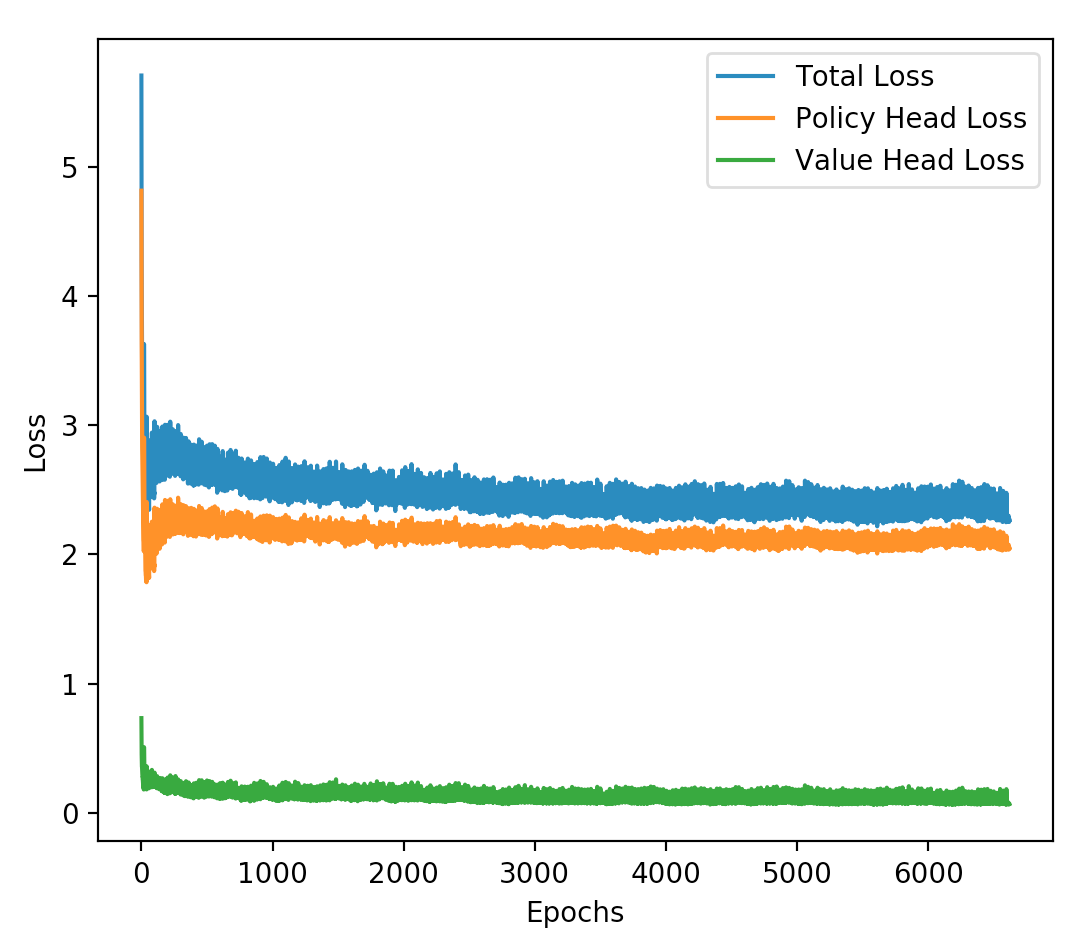}
    \caption{Loss history over epochs for training an agent from scratch through tabula rasa reinforcement learning.}
    \label{fig:fig11}
    %referred as figure 15
\end{figure}

\paragraph{Tabula rasa reinforcement learning} A model \(M_0\) with randomly initialized weights is trained for approximately 2 days using the training techniques and configurations as described in the previous sections, including time for generating training examples and training the model. The training loss history over epochs is illustrated in Figure 15: while it appears that the loss is slowly decreasing over time, it is evident that the learning was fairly inefficient. In addition, the loss value was volatile and show little signs of further decrease. Not surprisingly, when the trained model is matched against the best agent trained using the two-stage pipeline approach, it wins 0 out of 100 games. We argue that the failure is primarily due to two reasons. First, while the reward policy encourages the agent to move forward by rewarding the agent positively for more forward distance than its opponent, the reward policy does not however give the agent a consistent “win” state. In other words, so long as the current player has more forward distance, the win state can be drastically different while the value head of the network receives the same reward. Therefore, the gradients from both the value and policy head of the network can be ambiguous and not effective for training as the agent does not always receive consistent feedback. Secondly, the amount of training data used may be insufficient to train such a large model. Since the agent is trained starting from totally stochastic plays, the duration for a self-play game can be unbounded and hence the game is almost always terminated by a pre-set move limit, which, however, must be large enough so that a clear winner can be identified at the game’s terminal state. With these drawbacks, the efficiency of the data generation process is far from the that achieved by the two-stage training framework.  

\subsection{Experiments on Game Initialisation Techniques}
Another important aspect of the training framework is how different game initialisation techniques may affect self-play data generation, and thereby influencing the agent’s ability to adapt to diverse scenarios. As discussed in the previous section, two important game initialisation techniques that were deployed to introduce stochasticity to the data generation process include randomising the starting positions of all checkers and forcing the agent to take a fixed number of random moves at the beginning of each game. 

\paragraph{Randomising starting positions} This technique was initially proposed and implemented to address the lack of diversity of game-play patterns when generating heuristic-guided self-plays, as the number of unique move sequences are very limited if the agent always follows the same heuristic. However, through controlled experiments it was found that the effectiveness of this technique on the agent’s performance is relatively small.

For experiment, we trained two greedy agents, denoted by \(M_R\) and \(M_N\), on heuristic-guided self-play data, where \(M_R\) is trained entirely on games with random starting state while \(M_N\) is trained entirely on games with normal starting conditions. No other initialisation techniques were used. Starting from random models, both agents are trained for 100 epochs, and in each epoch 15,000 new games are generated and 5\% of training examples are retained for breaking correlation. All other training configurations follow those described in Table 1. 

By matching the two trained agents, we observed that the \(M_N\) is able to easily defeat \(M_R\) with an win rate of over 80\%. We also observed that \(M_R\) would often pick actions that are clearly locally sub-optimal, which is rather surprising as it was trained on a heuristic that always prefers locally optimal moves. We argue that this outcome is due to two factors:
\begin{enumerate}
    \item Most randomised starting states (and hence the subsequent states) are not representative of the distribution of the game states drawn from real matches.
    \item Since there is a large number of possible states, the performance of the agent depends heavily on what samples were generated during training.
\end{enumerate}

However, by training the agent on vastly different examples in each epoch, this technique has a positive impact of acting as a strong regulariser to the network, because when trained on such data the agent must extract more robust features from the board for making predictions instead of memorising or relying on a fixed pattern in the plays. This benefit is manifested by the fact that \(M_R\) is able to cope, though not necessarily optimally, with the highly likely unseen moves from its normatively trained opponent \(M_N\) during matches and win a certain proportion of them. 

For the above reasons, this technique is still deployed when generating 50\% of heuristic-guided self-play data. However, it is not used during MCTS-guided self-play reinforcement learning because the agent should be reinforced on real game-play data only. 

\paragraph{Initial random moves} Unlike randomising starting positions, forcing initial random moves of the agents can still result in game trajectories that are highly representative of the real game states distribution, but such initialisation can lead to much more diverse game-plays. 

For experiment, we trained two agents \(M_S\) and \(M_N\), where \(M_N\) is trained on a set of heuristic-guided self-plays where 50\% of the games started with the standard initial state while the other 50\% started with a random state. In contrast, \(M_S\) is also trained on a set of heuristic-guided self-plays, but 50\% of games started with a random state, 20\% of the games started with the normal initial state, and 30\% of games started with 3 random moves by each player. The reason for having 50\% of the training data to be drawn from games with a random starting state is to reduce the tendency for overfitting to initial starting strategies for both agents. All other training configurations remain the same as that described earlier. 

By matching the greedy agent with each of \(M_S\) and \(M_N\) where \(M_S\) and \(M_N\) are both guided by MCTS during the matches, we observed that \(M_S\) actually outperformed \(M_N\) as indicated by a 10\% increase in the win rate against the greedy agent, where \(M_S\) won 77/100 games against the greedy agent while \(M_N\) only won 67/100 against the same agent. We argue that this is because by introducing initial random moves, the agent is exposed to more starting strategies that the greedy agent would not have explored due to the apparent sub-optimality of these strategirs. However, by further increasing the percentage of games with random initial moves (while decreasing the percentage of games with normative starting states), we observe very little improvements in the agent’s performance. For these reasons, we retain 30\% of the self-plays to begin with random initial moves. The optimal number of initial random moves, however, remain an open area for further exploration, although throughout the experiments 3 random moves per player is observed to be a good default setting.

\subsection{Additional critical hyper-parameters}
\paragraph{Number of Monte Carlo simulations} Through controlled experiments of matching the agent against the greedy player, we observed that an increasing number of Monte Carlo simulations performed by the model would lead to a performance at least as good as the model allowed less simulations, as indicated by the win rate against the greedy player. However, we also observed that as the number of simulations becomes large (\(>\) 500), the marginal gain in performance decreases sharply while the compute resource required increases constantly. Another factor of consideration during our experiments was that performing a large number of simulations incurs heavy memory footprint; for example, since the Expansion stage of MCTS is performed in each simulation and on average there are 30 possible valid next actions, a single worker process running only 200 simulations will require storing more than 6,000 copies of game states in memory, which means around 72,000 copies of game states need to be cached with 12 independent workers. Through empirical analysis, 175 Monte Carlo simulations for each action was a good default for balancing the performance of and the resource required by the model.  

\paragraph{Tree search exploration constant \(c\)} The constant \(c\) in the MCTS selection stage controls the level of exploration of the agent during the search. Controlled experiments against the greedy agent indicates that this constant has a relatively small impact on the performance of our agent. However, a large \(c > 3.5\) tends to lead to an increase in performance (approx. 2\% increase in win rate) compared to the default choice of \(c = 1\), especially in the case when the number of simulations is large. The value \(c = 3.5\) is found by grid search to be a good default choice. While the effect of this constant is not always consistent, we do observe minor but consistent deterioration in the agent’s performance when a small number of Monte Carlo tree search \(< 100\) is combined with a large c \(> 3.5\). We argue that this is because as the agent is encouraged to explore other moves, it may not be able to search deeper into the game tree for planning non-local strategies.

\section{Conclusion and Future Work}
In this work, we have presented an approach for building a Chinese Checker agent that reaches the level of experienced human players with an effective combination of heuristics, Monte Carlo Tree Search (MCTS), and deep reinforcement learning. Through experimentation, we observed that the trained agent has learn common strategies played in Chinese Checkers and is robust to the dynamic states in the game. However, it remains an open topic worth further research that whether a Chinese Checker agent can be built tabula rasa to overcome some of the drawbacks of our agent as identified in this work, and whether expert-level multi-agents are possible in Chinese Checkers.

% placement []:
% p: page only contain table/fig, 
% h: here
% t/b: top/bottom of page

% \section*{References}

\bibliography{main}

\end{document}